# Generating Synthetic Human Blastocyst Images for In-Vitro Fertilization Blastocyst Grading

Pavan Narahari[1,2], Suraj Rajendran[1,2], Lorena Bori[3], Jonas E. Malmsten[4], Qiansheng Zhan[4], Zev Rosenwaks[4], Nikica Zaninovic[4], Iman Hajirasouliha[1,2,*]

[1]Institute for Computational Biomedicine, Department of Physiology and Biophysics, Weill Cornell Medicine of Cornell University, New York, NY, USA

[2]Caryl and Israel Englander Institute for Precision Medicine, The Meyer Cancer Center, Weill Cornell Medicine, New York, NY, USA

[3]IVIRMA Global Research Alliance, IVIRMA Valencia, Plaza de la Policía Local 3, 46015 Valencia, Spain

[4]The Ronald O. Perelman and Claudia Cohen Center for Reproductive Medicine, Weill Cornell Medicine, New York, NY, USA

[*]Corresponding:  imh2003@med.cornell.edu

## Abstract

The success of in vitro fertilization (IVF) at many clinics relies on the accurate morphological assessment of day 5 blastocysts, a process that is often subjective and inconsistent. While artificial intelligence can help standardize this evaluation, models require large, diverse, and balanced datasets, which are often unavailable due to data scarcity, natural class imbalance, and privacy constraints. Existing generative embryo models can mitigate these issues but face several limitations, such as poor image quality, small training datasets, non-robust evaluation, and lack of clinically relevant image generation for effective data augmentation. Here, we present the Diffusion Based Imaging Model for Artificial Blastocysts (DIA) framework, a set of latent diffusion models trained to generate high-fidelity, novel day 5 blastocyst images. Our models provide granular control by conditioning on Gardner-based morphological categories and z-axis focal depth. We rigorously evaluated the models using FID, a memorization metric, an embryologist Turing test, and three downstream classification tasks. Our results show that DIA models generate realistic images that embryologists could not reliably distinguish from real images. Most importantly, we demonstrated clear clinical value. Augmenting an imbalanced dataset with synthetic images significantly improved classification accuracy ($p < 0.05$). Also, adding synthetic images to an already large, balanced dataset yielded statistically significant performance gains, and synthetic data could replace up to 40% of real data in some cases without a statistically significant loss in accuracy. DIA provides a robust solution for mitigating data scarcity and class imbalance in embryo datasets. By generating novel, high-fidelity, and controllable synthetic images, our models can improve the performance, fairness, and standardization of AI embryo assessment tools.

## Introduction

In vitro fertilization (IVF) is a common treatment option for infertility issues, but IVF success rates remain relatively low with only ~30% of IVF users achieving a successful pregnancy and outcomes on the decline due to increasing female age[1]. Morphological assessment is critical for assessing the quality and selection of viable embryos. Typically, embryos are evaluated at day 5 between 108-112 hours post insemination (hpi) once they reach the blastocyst stage, however inconsistencies arise due to subjective grading by embryologists[2]. Grades are provided using systems like the Gardner grading system[3], which evaluates embryos on three morphological qualities: Expansion, Inner Cell Mass (ICM), and Trophectoderm (TE). Expansion is graded on a scale of 1-6, with 6 being highest quality and 1 being lowest quality, and ICM and TE are graded on a scale of A-C, with A representing high quality and C being low quality. The blastocyst score is a metric that combines the three scores and has been shown to correlate with ploidy status and implantation potential of an embryo[4].

Given the importance in accurate grading and its correlation with selecting viable embryos, recent advancements in artificial intelligence (AI) have shown promise in standardizing this embryo selection process[5,6,7,8,9,10]. A primary challenge, however, is that these deep learning models require thousands or even millions of images to train effectively and the quality of these AI models is heavily reliant on both the quantity and quality of the available embryo data. Furthermore, real-world clinical datasets are rarely balanced and often suffer from sharp imbalances across scoring categories. Training on such data can lead to biased models that perform poorly on under-represented scores and fail to learn the full variability among embryos. This data scarcity problem is compounded by the ethical and privacy concerns that restrict the sharing of large datasets between clinics. Consequently, clinics with smaller datasets may be unable to implement robust classification models, reinforcing their reliance on inconsistent manual evaluation.

Synthetic data generation offers a powerful solution to challenges of data scarcity and imbalances. Generative models such as generative adversarial networks (GANs) and diffusion models[11] have been widely used to augment limited datasets in both medical[12] and non-medical applications[13]. Within the embryology space, several generative models exist but each come with their own shortcomings. While earlier studies employed GAN-based approaches[14,15], subsequent evaluations have shown that diffusion models produce superior image quality[16]. A few studies[17,18] have applied diffusion models, but are constrained in their clinical utility. The Presacan et al. study[17] allows for generation of images at various embryo stages, which is useful for morphokinetic tracking and determining when embryos have become blastocysts, but not directly applicable to selection of the most viable embryo. Additionally, they do not present memorization metrics on their synthetic images while using a 1000 image training dataset. This is an important consideration working with medical data, as extensive research has shown that diffusion models tend to memorize training data, especially when datasets are small[19,20,21]. The Golfe et al. study[18] allows for conditional generation of oocytes depending on whether it turns into a blastocyst, a task that similarly does not directly aid in selecting specific embryos for transfer during IVF. Their FID scores are also much higher than existing generative models for embryology showing the synthetic images may not be as perceptually similar to real images.

We present DIA, a framework for generating synthetic day 5 embryo images that match the quality and variability of imaging data captured via Embryoscope+ imaging systems. A key objective was to provide users with granular control over the output. To achieve this, our models are conditioned on both Gardner-based morphological scoring categories and the z-axis focal depth (FD) between -75 and +75. From this framework, we developed four models. Three are designed to generate images based on a single Gardner category: DIA-E (Expansion), DIA-I (ICM), and DIA-T (TE). The fourth model, DIA-EIT, generates embryos with a specific cumulative Expansion-ICM-TE score. The individual models (DIA-E,

-I, -T) are intended to augment datasets for downstream classification tasks, whereas the DIA-EIT model is an educational tool for embryologists to visualize embryos of a specific cumulative score.

We employed a three-pronged evaluation: Fréchet Inception Distance (FID) to measure image quality, the Self-Supervised Descriptor for Image Copy Detection (SSCD)[22] to quantify memorization, and human-based Turing tests to assess perceived realism. We demonstrate that augmenting datasets with our synthetic images leads to improved performance in downstream classification models tasked with scoring real embryos, whether the original datasets are limited, imbalanced, or even already large. We show that we generate realistic and novel images that can be used to augment embryo datasets, which can improve the standardization of embryo quality evaluation and lead to better IVF outcomes for patients.

## Results

### Model Development

The overall workflow for DIA model development and evaluation is shown in Figure 1. Real images were initially used to train an autoencoder that was subsequently used during DIA model training. Synthetic images were evaluated for similarity to real images using FID, memorization using SSCD, and their realism through a Turing Test with embryologists. To evaluate performance on downstream classification models, synthetically generated images from DIA-E, DIA-I, and DIA-T were evaluated on three separate tasks against a set of baseline classification models trained on only real images.

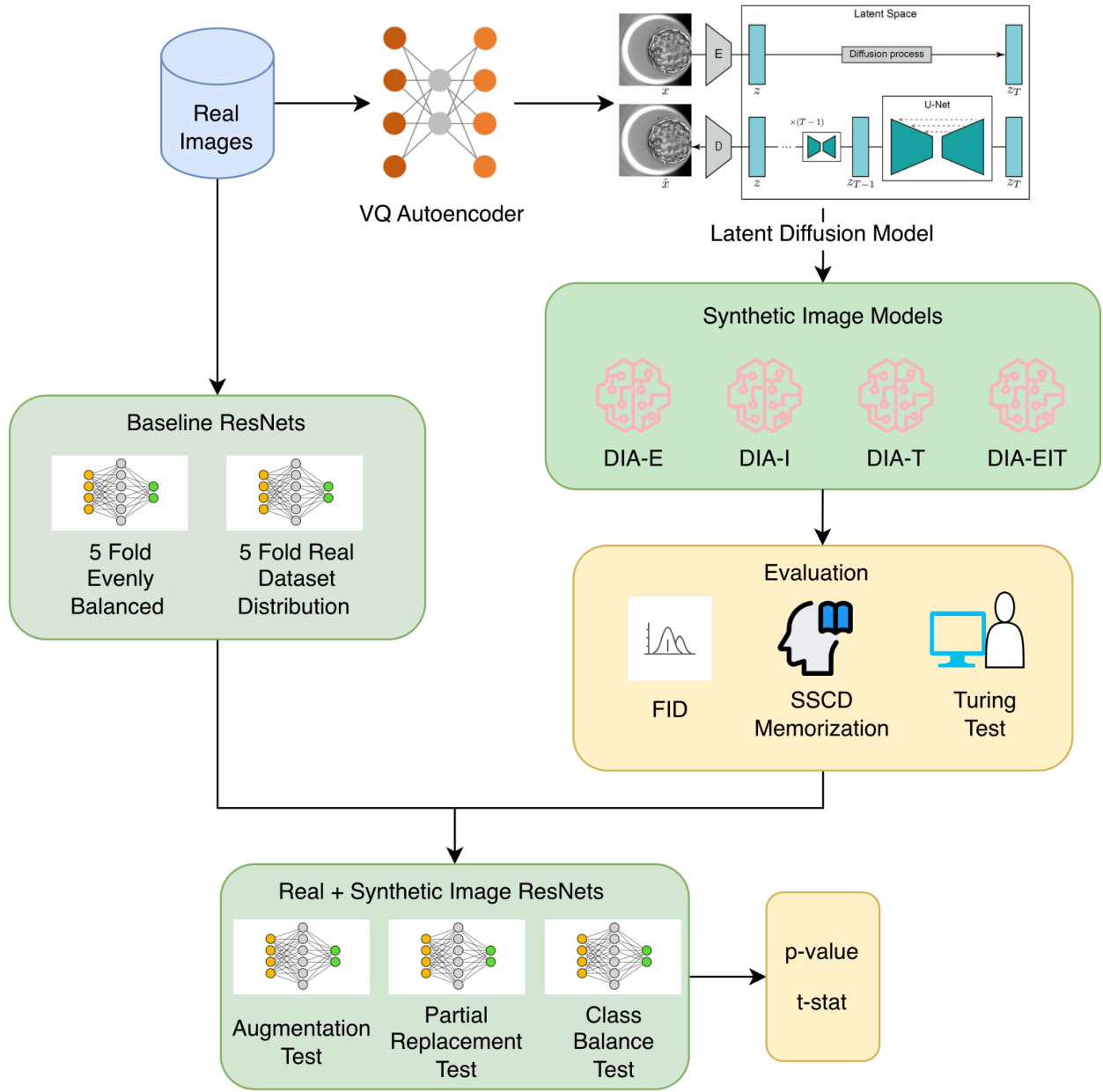

**Figure 1: Study Workflow, Model Development, and Evaluation.** Real images were used to train a VQ-VAE. The autoencoder was used as part of LDM training, which led to DIA models. DIA was evaluated on quantitative and qualitative metrics using FID, SSCD, and a Turing Test. We evaluated downstream classification performance by comparing baseline models trained only on real images against augmented models trained on combinations of real and synthetic images across three distinct experimental setups: an Augmentation Test, a Partial Replacement Test, and a Class Balance Test.

**Latent Diffusion Models**

We trained four latent diffusion models[23] (LDM) for synthesizing day 5 embryo images (108–112 hpi). The models were designed to enable conditional generation based on morphological scores. DIA-E, DIA-I, and DIA-T each accept conditioning information for a single scoring category (Expansion, ICM, or TE) along with FD value. DIA-EIT accepts all three scoring parameters plus FD, allowing full control over the generated embryo's Gardner morphological qualities.

Conditional control was achieved using classifier-free guidance[24] (CFG) during training. During inference, a CFG-scale value determines how closely generated images adhere to the conditioning input, with higher scales enforcing stronger adherence at the potential cost of increased artifacts. As part of model evaluation, we tested three CFG-scales during image generation: 2.5, 5.0, and 7.5.

Images for model development were obtained at Weill Cornell Medicine (WCM) during 2021–2022 using an Embryoscope+. Images were captured between 108 - 112 hpi, but expanded to 112-120 hpi for less represented scoring combinations, meaning synthetic images should be produced with features representing day 5 embryos. These time points were chosen due to their clinical relevance for embryo selection. Embryo grades were annotated by WCM embryologists following the modified Gardner grading system. Each morphological category was converted to a numerical scale as shown in Supplemental Table 1 with resulting values for Expansion being between 1-4 and ICM/TE between 1-3, where 1 corresponds to the highest quality. Representative images from each model at varying scores and FD of 0 are compared to real embryo images in Figure 2.

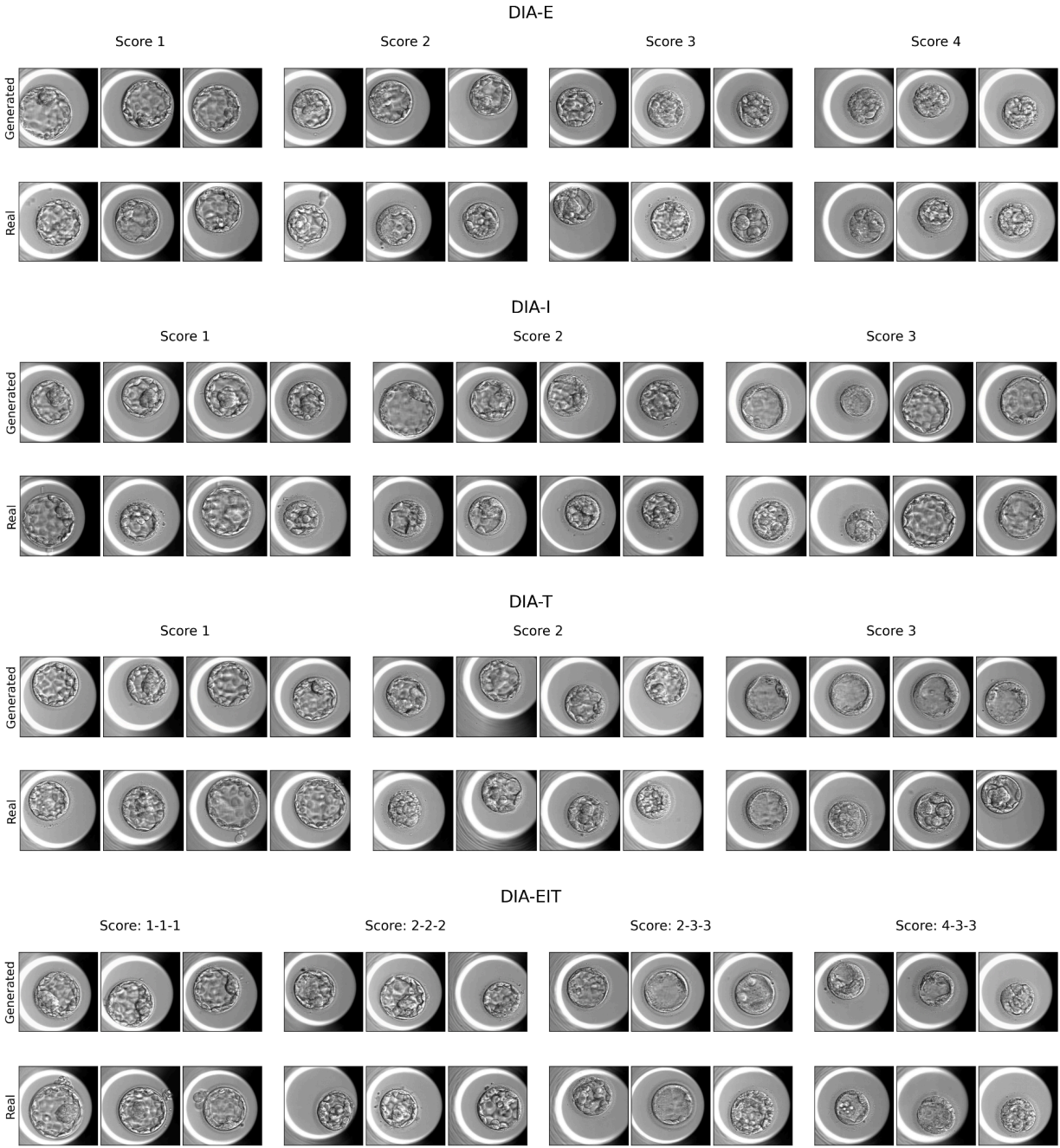

**Figure 2:** Sample images for various scores for each model compared to real images. The top row shows generated images from DIA whereas the bottom row shows real images. For DIA-EIT cumulative scores are written with format Expansion-ICM-TE.

## Quantitative Evaluation of Synthetic Images

To comprehensively evaluate our models, we used a combination of metrics to assess the generated images. We employed FID for perceptual quality and the SSCD to measure memorization.

We also analyzed the impact of the CFG-scale hyperparameter during image generation. Consistent with prior work, higher CFG-scales tended to increase FID scores[24], though in some cases improving downstream classification performance, showing there may be better perceptual similarities to real images at higher scales. The final version for each model was ultimately selected based on a balance of both FID and SSCD scores.

**FID**

We used FID scores to evaluate the resemblance of synthetic images to the real image distribution. We calculated all FID scores against a separate 1000-image test dataset, ensuring no overlap with the training data. This test set was also evenly balanced across cumulative score combinations, similar to the training set, to maintain a consistent representation of the embryo distribution. For each of our models, FID scores were calculated by comparing 1000 generated images against the test set. This process was repeated at three different CFG-scales.

To establish a dataset-specific benchmark, we calculated the FID between two independent real-world embryo datasets, obtaining a score of 14.71. This score represents the natural variance in the data and serves as our practical target. Prior work in generative embryology[15,17,18] reports lowest scores from 10–94.32. However, we note that the lower end of this range may be artificially optimistic, as some methods calculate FID against the training data, which can favor memorization[19]. Table 1 shows the FID values for each model at the three different CFG-scales used during inference of the generated image sets.

| **CFG-SCALE** | **DIA-E** | **DIA-I** | **DIA-T** | **DIA-EIT** |
|---|---|---|---|---|
| 2.5 | 22.2 | 21.68 | 23.43 | 23.63 |
| 5.0 | 23.02 | 23.22 | 27.89 | 24.62 |
| 7.5 | 25.34 | 26.34 | 34.71 | 30.42 |

**Table 1:** FID scores for all models at various CFG-scales

**Memorization**
While FID scores are the standard metric for evaluating generative models, FID values can be deceptively lower by memorization of the training data. Therefore, we selected each model based on a combination of both FID and memorization metrics. To quantify memorization, we used SSCD to measure the similarity between generated images and the training set. SSCD works by extracting a feature vector from a generated image and computing its cosine similarity against the feature vectors of all images in the training set. The training image with the highest similarity is considered the closest match. As discussed in the Methods, we empirically determined a threshold of >0.85 to indicate high similarity and potential memorization.

For each model and CFG-scale we randomly selected 50 generated images and plotted a histogram of their SSCD scores against the training data. Figure 3 shows the histograms for each model and the three images with the highest SSCD score. The distributions suggest that the models did not memorize the training data and are instead generating novel embryo images.

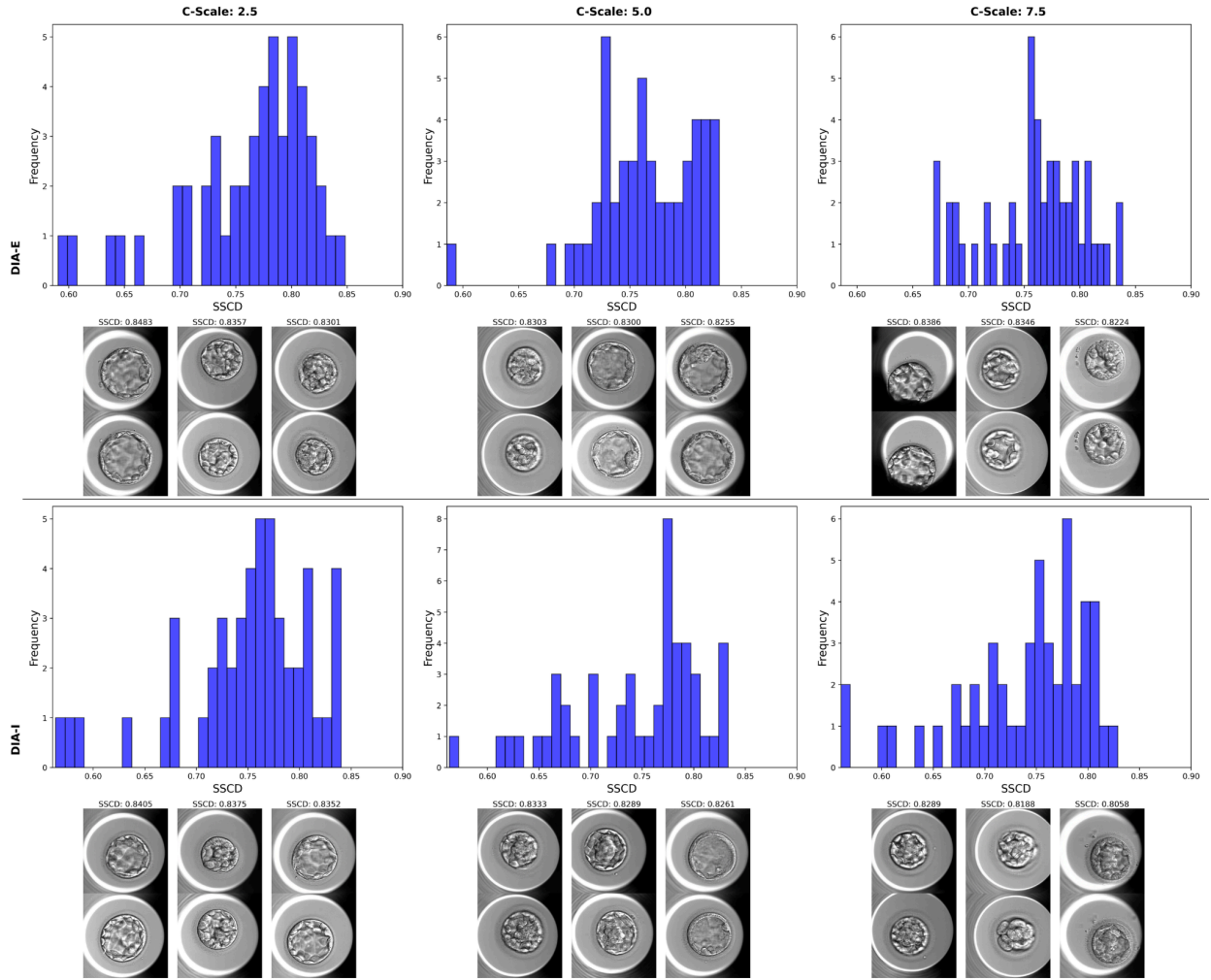

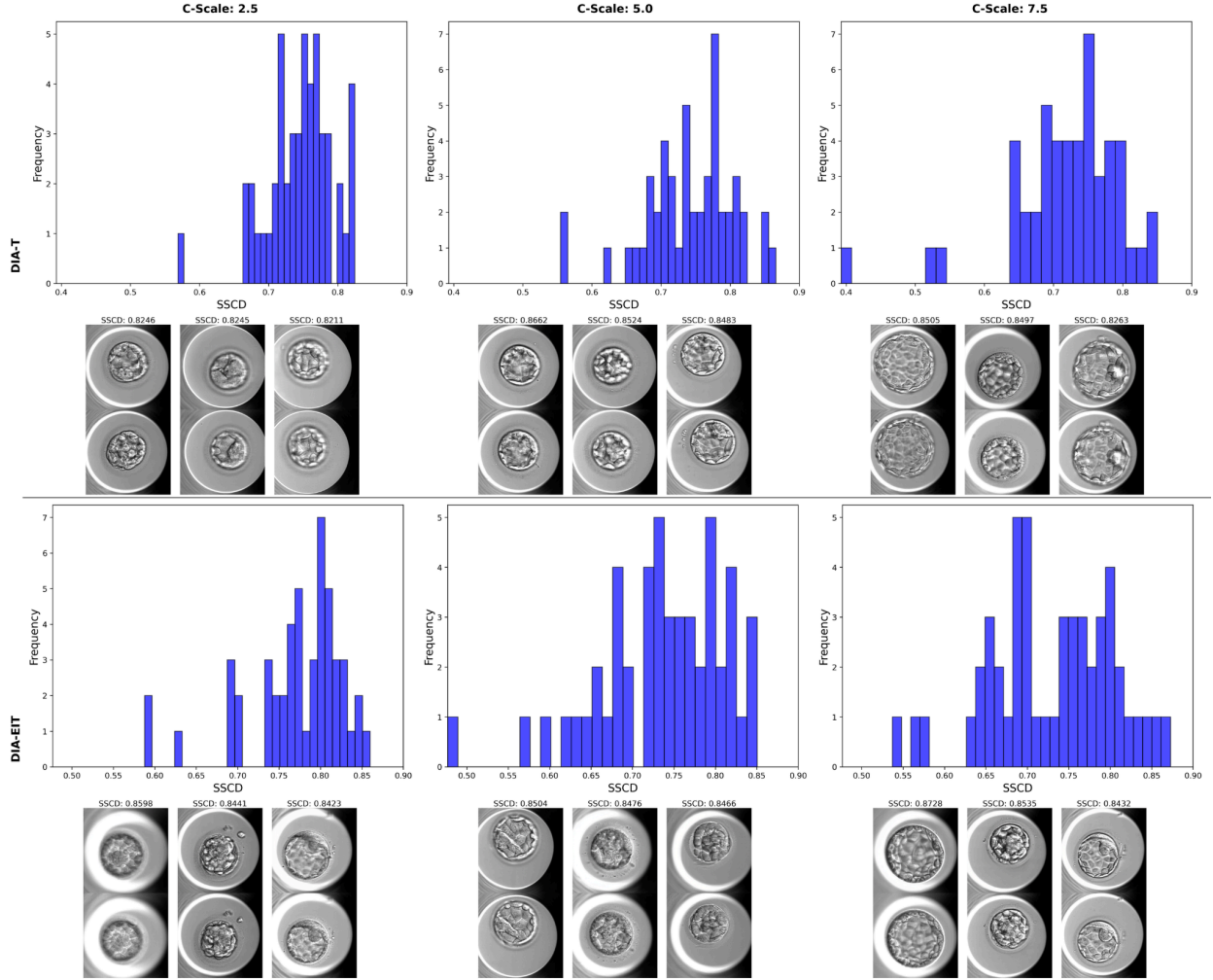

**Figure 3:** SSCD histogram scores for each model computed on 50 randomly selected generated images compared against the training dataset at three different CFG-scales. Below each histogram, the top three generated images with the highest SSCD scores are shown (top row), along with their closest matching real images from the training set (bottom row).

## Qualitative Evaluation of Synthetic Images

### Turing Test

We conducted a Turing test in which four expert embryologists assessed the realism and morphological accuracy of the generated images. The results are summarized for each model in Figure 4 as averaged count confusion matrices. During the Turing test, embryologists were presented with one image at a time and asked to perform two tasks: first, classify the image as real or synthetic, and second, assign it a quality score. This methodology produced two types of confusion matrices. For the real vs. synthetic matrices, a weak or indistinct diagonal was the desired outcome, as this would indicate that embryologists could not reliably distinguish synthetic images from real ones. In contrast, for the score-based matrices, a strong diagonal was expected, which would reflect high agreement between the embryologists' assigned scores and the conditioning scores. Our results aligned with both of these expectations.

Precision, recall, and accuracy metrics are shown in Supplemental Table 2 for each model, separated by real and synthetic ground truth. Real vs Synthetic judgement accuracy was 64.3% for DIA-E, 58.6% for DIA-I , 59.1% DIA-T, and 61.6% for DIA-EIT, indicating performance only slightly above random guessing. Scoring accuracy on synthetic images was 67.7% for Expansion (DIA-E), 71.9% for ICM (DIA-I), 74.0% for TE (DIA-T), 49.3% for Expansion (DIA-EIT), 60.4% for ICM (DIA-EIT), and 70.8% for TE (DIA-EIT). Interestingly, across all models and scoring categories, embryologists were more accurate when evaluating synthetic images than real ones, suggesting higher variability in the judgement of real embryos, while synthetic images exhibited clearer features associated with their conditioning scores.

We also performed a Cohen's Kappa Agreement calculation between embryologists as shown in Supplemental Table 3. We saw low agreement between embryologists on real vs synthetic judgement and higher agreement on score values. Interestingly, we saw embryologists in agreement more often on synthetic scores versus real image scores, reinforcing that generated images are clear, consistent representations of a scoring category.

Finally, we recruited non-embryologists with AI and computational backgrounds to participate in the Turing test, assessing whether their familiarity with generative-model artifacts improved their ability to distinguish synthetic images. As shown in Supplemental Figure 1, these participants also performed near chance, further confirming the realism of generated images.

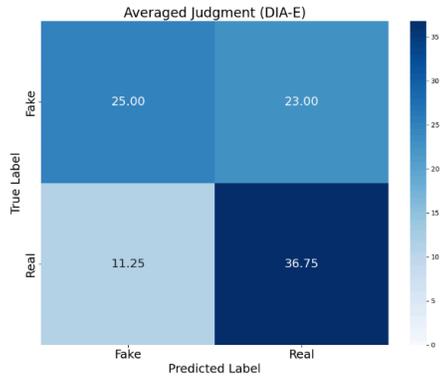 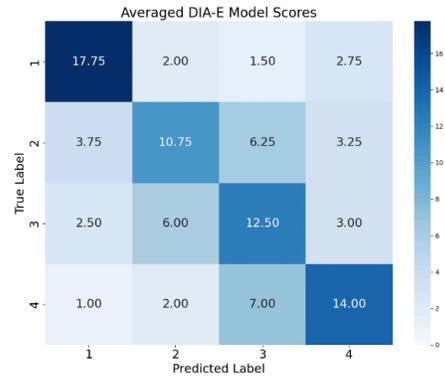

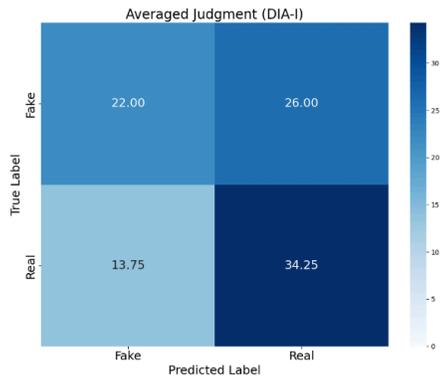 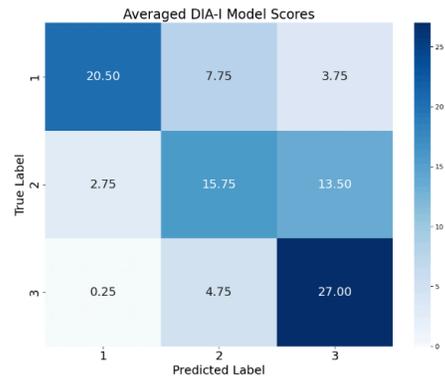

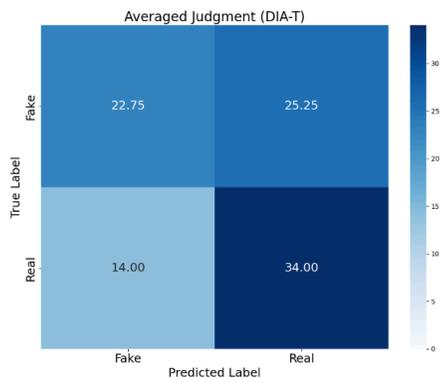 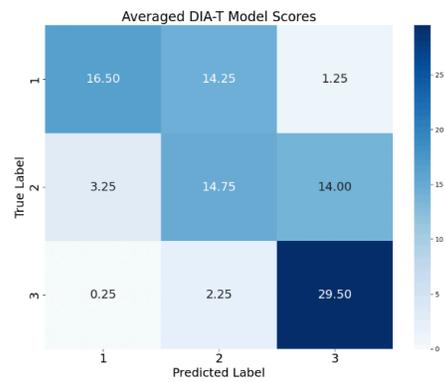

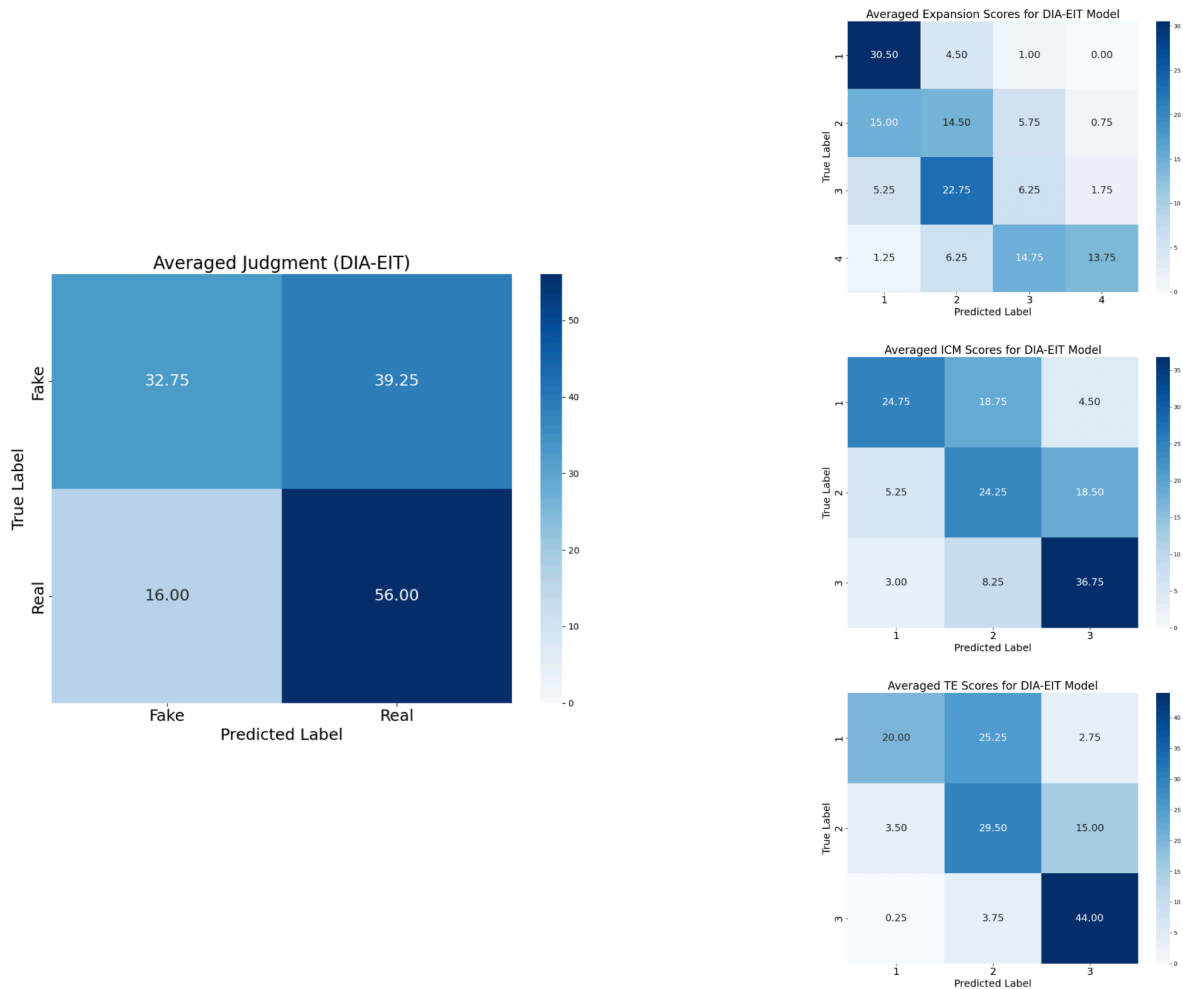

**Figure 4:** Confusion matrices from the Turing test averaged across four embryologists. The matrices on the left show the classification of real versus synthetic images. The matrices on the right show predictions of embryo quality scores compared to ground truth for the relevant scoring categories of each model.

**Conditioning Information**

To qualitatively validate that conditioning information that was properly captured during image generation, we performed a simple test where a fixed Gaussian noise sample was passed through each model while varying only the conditioning score or the FD value. Supplemental Figures 2, 3, and 4 show the resulting images for DIA-E, DIA-I, DIA-T. Each figure displays a clear and continuous gradient corresponding to changes in the respective morphological features, as well as along the FD values, demonstrating that the models clearly learned difference in scores and FD values.

**Downstream Tasks**

Given the downstream goal of our models is to augment datasets for classification tasks, we performed several experiments using ResNet classification models.

**Baseline ResNets**

We trained several real image only baseline ResNet classification models. These models serve as the control group for our downstream experiments, establishing the performance benchmarks against which synthetic data augmentation is measured.

Given that classification models can achieve better performance when focusing on a specific scoring category rather than cumulative scores, we trained individual classification models for each of the three Gardner categories: Expansion, ICM, and TE. To facilitate statistical significance testing in downstream tasks, we established two baselines, both trained using a five-fold cross-validation structure for each scoring category:

1. Real-World Distribution Dataset**:** This baseline reflects the unbalanced, real-world distribution of embryo scores, with percentages for each score derived from the distribution in the embryos annotated in 2022 at WCM.
2. Evenly Balanced Dataset**:** This baseline consists of 4000 training images and 1000 validation images roughly balanced across scores for each Gardner category and across cumulative score combinations.

**Class Balance Test**

In most real-world datasets, embryo images are unevenly distributed across score categories. Training directly on such imbalanced data can bias models toward overrepresented scores, while aggressive downsampling can reduce the diversity of examples within each class. To address this, we compared Baseline 1 to those trained on datasets augmented and balanced using synthetic images.

Each augmented dataset contained a total of 4000 training images, with synthetic images added to balance the score distributions. We then compared the validation accuracies of ResNets trained on real-only data versus these augmented datasets, running experiments across three different CFG-scales. As shown in Figure 5a, augmentation with synthetic images led to significant ($p < 0.05$) improvements in validation accuracy across nearly all models and CFG-scale conditions. The sole exception was the Expansion model using DIA-E images at a CFG-scale of 5.0. Statistical significance was determined using p-values calculated from the five-fold validation accuracies for each model. Notably, classification accuracy also improved for underrepresented score categories (Supplemental Figure 5). Individual accuracies for each experiment are detailed in Supplemental Table 4.

**Augmentation Test**

For larger datasets such as WCM, which can already be balanced across score categories, we sought to determine whether adding synthetic images could further improve performance when training classification models. Synthetic images were added at three different CFG-scales to ResNets trained for Baseline 2.

Synthetic image counts were added in increments of ~4000 images till a majority of the CFG-scale experiments led to a reduction in t-stat compared to the previous increment's t-stats. Figure 5b. shows averaged validation accuracies between the real-only and augmented ResNets. All models and CFG-scales demonstrated statistically significant improvements in validation accuracy. These results suggest that even in initially large real-only datasets, adding synthetic images can improve performance of classification models. Individual accuracies for each experiment are in Supplemental Table 4.

**Partial Replacement Test**

We next tested how replacing real images with synthetic images affected model performance to determine the threshold at which removing real data led to a statistically significant decrease in ResNet accuracy. This experiment simulates the extent to which synthetic images can substitute for real images while maintaining equivalent performance. Using the same balanced real image baseline ResNets as the previous experiment (Baseline 2), we gradually replaced portions of the real training images with synthetic ones.

Each removed real image was replaced with a synthetic image generated with the same score and FD. Replacement tests were performed at 20%, 40%, 60%, and 80% substitution levels using images from each diffusion model at three different CFG-scales. ResNet validation accuracy was then compared to Baseline 2 models, and p-values were calculated across folds. A p-value greater than 0.05 indicated that the performance decrease was not statistically significant.

We observed that replacing real images with synthetic ones led to performance declines of varying degrees across models as shown in Figure 5c. At 20% replacement, all three CFG-scales of the Expansion model showed statistically significant performance decreases, while in contrast all three CFG-scales of the ICM model and two of the three CFG-scales of the TE model maintained non-significant decreases. At 40% replacement two of three ICM scales and one of three TE scales remained non-significant. Beyond 60% replacement, all models and CFG-scales exhibited statistically significant performance declines, indicating that while synthetic images can supplement real data, they cannot fully replace it without loss in predictive performance. Individual accuracies for each experiment are in Supplemental Table 4.

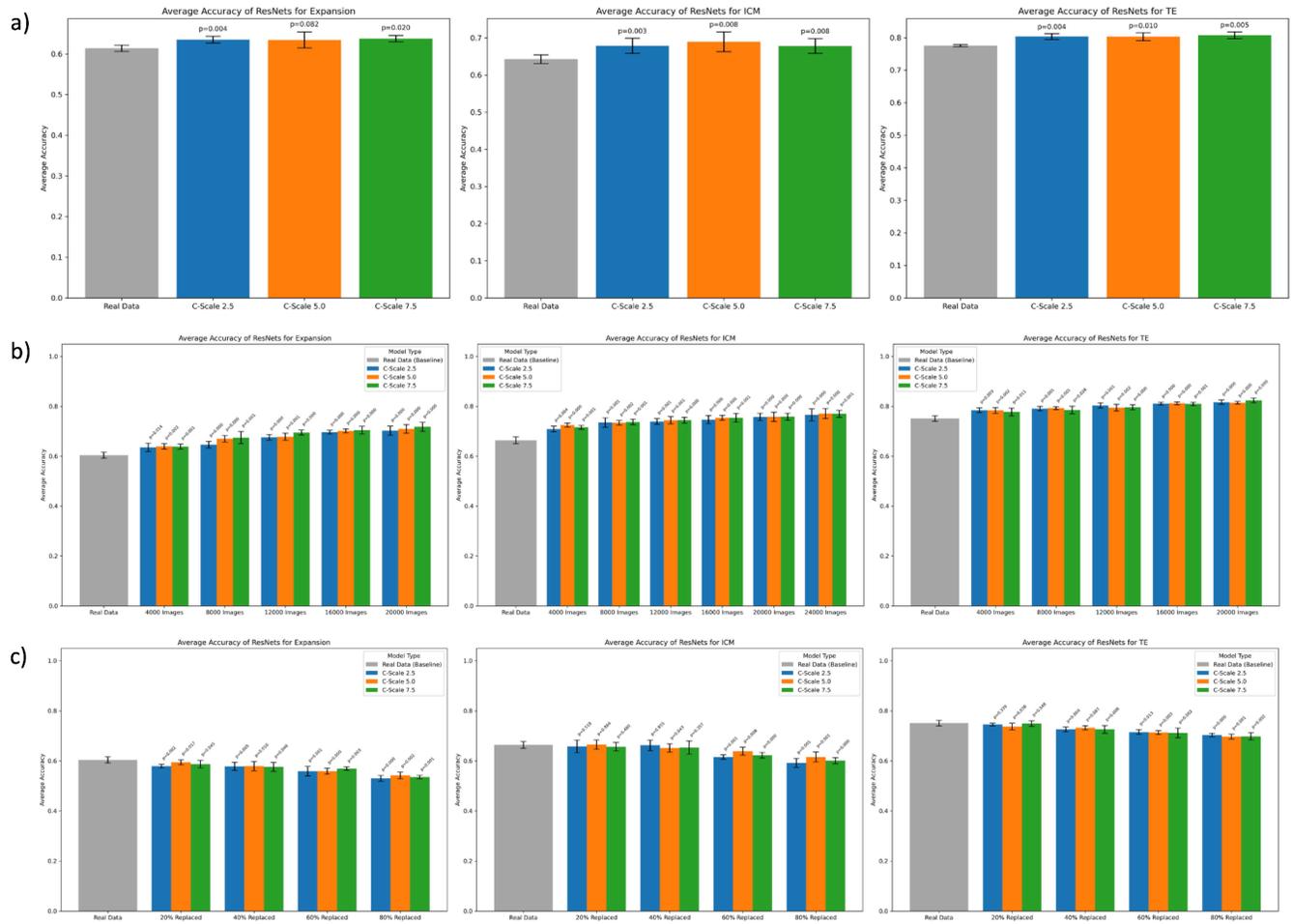

**Figure 5: a)** Average five-fold accuracy for ResNet models trained on the baseline real image dataset following a natural real-world distribution, compared to models trained on a balanced dataset with synthetic images. **b)** Average five-fold accuracy for ResNet models trained on the baseline balanced real dataset compared to those augmented with synthetic images. **c)** Average five-fold accuracy for ResNet models trained on the balanced real dataset compared to models where varying percentages of real images were replaced with synthetic images.

## Discussion

In this study, we introduced the DIA framework, which includes four models (DIA-E, DIA-I, DIA-T, and DIA-EIT) capable of generating novel, high-fidelity synthetic images of day 5 human embryos. We demonstrated that these models provide granular control over clinically relevant morphological features, Expansion, ICM, TE, and FD, addressing a key gap in prior generative models for embryology.

Our quantitative and qualitative evaluations confirm the models produce images that are both realistic and novel. The DIA models achieved low FID scores against a separate, balanced test set, indicating a high degree of similarity to the real image distribution. We established a domain-specific SSCD threshold of >0.85 to rigorously assess novelty, confirming that our models are not merely reproducing the training data, a methodological step missing from most prior work in this field.

Furthermore, our visual Turing test revealed that expert embryologists could not reliably distinguish synthetic images from real ones, with real vs synthetic judgment accuracy not much better than chance (e.g., 0.586 to 0.643) highlighting the perceptual realism of generated images. Interestingly, embryologists showed higher agreement and accuracy when grading synthetic images compared to real ones. This suggests our models generate clear, consistent representations of each scoring category.

The primary contribution of this work lies in its demonstrated clinical utility for downstream classification tasks. We systematically evaluated the impact of synthetic data augmentation in three distinct, clinically relevant scenarios using five-fold cross-validation and statistical testing. In our class balance test, augmenting a real-world, imbalanced dataset with synthetic images led to statistically significant improvements in classifier validation accuracy across nearly all models. Notably, this augmentation also improved performance on the under-represented scores. This directly addresses a major challenge for clinics, where training on imbalanced data can lead to biased models. In the augmentation test, we found that even for a large, already balanced dataset, adding synthetic images provided statistically significant performance gains. This finding is particularly powerful, as it shows that synthetic data can be an asset even for high-resource clinics, not just those with limited data. Finally, the partial replacement test demonstrated that synthetic images can replace a portion of real data (up to 40% for some models) without a statistically significant drop in performance. However, performance declined significantly at higher replacement levels, indicating that while synthetic images are a powerful supplement, they cannot fully replace the diversity and complexity of real data.

Beyond data augmentation, the DIA-EIT model, with its full granular control over all three Gardner categories, shows promise as a valuable educational tool. Through discussions with embryologists, we determined it could be used to train junior embryologists by allowing them to visualize and study specific and rare combinations of morphological features on demand.

Our work builds upon previous efforts in generative modeling for embryo imaging but makes several key distinctions. Unlike prior models that used GANs, which have been shown to produce lower-quality images, we used diffusion models. We also address several limitations of previous generative embryo papers. First, many models lack granular conditioning, focusing on broad developmental stages rather than clinically-relevant morphological scores used for embryo selection. Our framework provides this precise control. Second, our evaluation is more rigorous. We compute FID scores against a dedicated test set, whereas other papers report scores against their training data, which favors memorization and can overinflate reported FID scores. Third, we quantitatively evaluate for memorization, a critical step often omitted, ensuring our models generate novel images. Finally, we provide statistically significant evidence for downstream task improvement to demonstrate clinical utility.

Despite these strengths, our study has limitations. The models were trained on a large dataset from a single, high-resource institution using an Embryoscope+ time-lapse-imaging (TLI) system and annotated with Gardner grading system. Consequently, the synthetic images may not generalize directly to other clinics using different imaging technology or grading systems, potentially requiring clinic-specific adaptations to be useful. The models also rely on embryologist annotations, which carry inherent inter-observer variability. Our models are also limited to a specific, although clinically critical, time window (primarily 108-112 HPI) and cannot model the full developmental cycle. The conditional control is also limited to the three Gardner categories and FD, excluding other potential morphological factors.

Future work should focus on addressing these limitations. Training models on more diverse datasets from multiple clinics and imaging systems would enhance generalization. A significant advancement would be to incorporate temporal conditioning, enabling the generation of entire time-lapse videos of embryo development from fertilization to the blastocyst stage. Finally, expanding conditional control to include other biological or morphological features could further increase the utility of synthetic embryo generation.

In conclusion, the DIA models represent a significant step forward in generative modeling for reproductive medicine. We have demonstrated the creation of realistic, novel, and controllable synthetic embryo images and have rigorously proven their ability to improve downstream classification models. This work provides a tangible path toward mitigating data scarcity and imbalance, ultimately aiding in the standardization of embryo assessment and supporting the goal of improving IVF outcomes for patients.

## Methods

The study was performed in accordance with relevant guidelines and regulations. The study was approved by the Institutional Review Board at Weill Cornell Medicine (numbers 1401014735 and 19–06020306). IRB determined that this research meets the exemption requirements at HHS 45 CFR 46.104(d) and is secondary research for which consent is not required. A waiver of informed consent was granted from the IRB as the images were de-identified for this retrospective review of clinical data. The embryo imaging was performed as part of the standard care procedure during the preimplantation and IVF cycle. No discarded embryos were used. In this study, information, which may include information about biospecimens, is recorded by the investigator in such a manner that the identity of the human subjects cannot readily be ascertained directly or through identifiers linked to the subjects. Moreover, the investigators do not contact the subjects, and the investigators will not re-identify the subjects. As such, informed consent was not obtained, and participants did not receive compensation for the study.

### Data Collection and Grading

Embryo images were captured every 20 minutes using the EmbryoScope+ TLI system, which acquires images at multiple FDs throughout embryo development. Day 5 is a clinically relevant time point for embryo development where they reach the blastocyst stage and are graded for transfer. At WCM, embryos are graded using the Gardner scale typically between 108-112 hours by embryologists, where they assess three primary morphological features: Expansion, ICM, and TE.

For the purposes of this work, embryologist annotated Gardner grades were converted into simplified numerical scores, closely aligned with the WCM blastocyst scoring system, which has been shown to correlate with ploidy and implantation potential. Day 5 grading plays a critical role in determining embryo quality and selection for transfer.

### Vector Quantized Variational Autoencoder

A custom VQ-VAE was trained to encode embryo images into a latent representation for use in downstream LDMs and to decode them latent images back into pixel space. The model utilizes a convolutional encoder-decoder architecture with residual blocks at multiple resolution levels, compressing 256x256 pixel input images into a latent space 4 times smaller than the original resolution.

Vector quantization was achieved by mapping the continuous encoder outputs to the nearest entry in a discrete codebook of 8,192 learned embeddings to produce a 3 dimensional vector. The VQ-VAE was trained on a dataset of 20,199 embryo images of FD 0 from the WCM 2021 dataset, with 2,020 images reserved for validation. A complex loss function, VQLPIPSWithDiscriminator, was used, which combines perceptual similarity metrics with an adversarial discriminator to optimize reconstruction quality. Training was performed with a batch size of 8 and gradient accumulation over 2 steps, continuing until the validation loss converged. The VQ-VAE model was used to encode and decode images for subsequent LDM training.

## LDMs

### Dataset

All LDM training images were obtained from WCM-annotated embryo datasets. Embryos were categorized according to their day 5 (108–112 hpi) annotations for Expansion, ICM, and TE quality. Rather than balancing individual scoring categories independently (e.g., only Expansion), we constructed datasets balanced as much as possible across cumulative scoring combinations, ensuring that each model was exposed to the full context of variability present within the relevant score while varying the other categories. Datasets were also balanced across FDs to ensure consistent representation of focal planes.

The training set for DIA comprised 85716 WCM images from 2022. For each cumulative score combination, we included up to 250 images at each FD. Since certain biological combinations of scores are rare, some categories contained fewer examples. In these cases, we extended the temporal window to include additional images captured between 112–120 hpi.

The validation and test sets consisted of 8716 and 1000 WCM 2021 images, respectively. These datasets were similarly balanced across cumulative scoring combinations and FDs. Of the 36 possible cumulative scoring combinations, four were absent from the validation set and seven from the test set. No overlap existed between the training, validation, and test sets. The dataset distributions across scores and FDs are detailed in Supplemental Figures 6, 7, and 8.

### LDM Architecture and Training

All input images were resized from 800x800 to 256x256 for computational efficiency. The LDMs were trained using the latent representations generated by the custom-trained VQ-VAE described in the previous section. During training, Gaussian noise was progressively added to these latent representations, and a UNet-based denoising network was trained to predict and remove the noise at each timestep. This reverse process effectively learns to reconstruct clean latent embeddings, thereby capturing the data distribution of embryo images.

Conditional information was incorporated using a multiclass embedding module that independently encoded the Expansion, ICM, TE, and FD categories into 512-dimensional embeddings. These embeddings were concatenated and integrated into the UNet architecture via cross-attention layers. CFG was implemented by randomly omitting conditioning information from 10% of training samples, enabling the model to learn both conditional and unconditional generation.

Training was performed using 500 diffusion steps, a base learning rate of 0.000001, and a linear noise schedule between 0.0015 and 0.0195. Model performance was monitored on the validation set using an exponential moving average (EMA) of the reconstruction loss. The three checkpoints with the lowest validation losses were retained for downstream evaluation and image generation to choose the final model version. Training continued until at least 15 epochs passed without improvement in the top three validation loss checkpoints.

**Image Generation from DIA**

Image generation from the trained diffusion models was performed using deterministic DDIM[25] sampling to accelerate inference. Images were generated using 50 DDIM timesteps with a stochasticity parameter of 1.0. During sampling, conditioning information corresponding to relevant embryo scores and FD was provided along with a CFG-scale to steer image generation toward the conditioning signal. Three CFG-scale values (2.5, 5.0, and 7.5) were evaluated for all models.

**Synthetic Image Evaluation**

**FID**

Fréchet Inception Distance (FID) was used to quantify the similarity between the distributions of generated and real embryo images. FID was calculated by passing each image set through a Inception-v3 network pretrained on ImageNet to extract feature representations. The mean and covariance of these feature vectors were then compared between the image sets to compute FID. Lower FID scores indicate closer alignment, with a score of 0 representing identical feature distributions. Approximately 1000 images were used for each FID calculation. As a baseline, we computed the FID between two independent real-world embryo datasets, obtaining a score of 14.71, which served as a rough lower bound for generated versus real image datasets.

Synthetic images for FID calculations were evenly distributed across FDs and scores. For example, DIA-E models produced ~250 images per score with balanced FDs between -75 and +75. FID scores were reported separately for the three tested CFG-scales and compared against the WCM 2021 test set.

**Memorization**

To mitigate concerns of memorization, we evaluated each model and CFG-scale by randomly selecting 50 generated images and identifying their closest real image in the training set using SSCD (Self-Supervised Descriptor for Image Copy Detection). SSCD extracts feature representations from images and computes the cosine similarity between each generated image and all training images. The training image with the highest similarity is considered the closest.

Previous studies using diverse datasets such as Imagenette and LAION have suggested a threshold of >0.6[26] to indicate memorization. However, because our dataset consists entirely of embryo images, there is a high inherent similarity between images. We therefore conducted an empirical analysis to determine an appropriate threshold. This involved augmenting real images via transformations to observe resulting SSCD scores against the training dataset and examining generated images at various SSCD values alongside their closest real counterparts. Based on this analysis, we set a threshold of >0.85 to indicate high similarity and potential memorization (see Supplemental Figures 9 and 10).

**Turing Test**

To assess the realism and morphological accuracy of generated images and score conditioning, we developed a web portal to conduct a Turing test as shown in Supplemental Figure 11. Users were presented images one at a time and evaluated whether they thought the image was real or fake and the score they would provide the embryo. For the three single-score models (DIA-E, DIA-I, and DIA-T), we evaluated 96 images each (48 real, 48 synthetic). These images were evenly split across the models' respective score ranges (1-4 for DIA-E; 1-3 for DIA-I/T). For the cumulative-score model (DIA-EIT), we evaluated 144 images (72 real, 72 synthetic), which were evenly split across all 36 possible score combinations for Expansion, ICM, and TE. In total, 432 images were evaluated. To eliminate bias, all images were drawn randomly from the test sets rather than being hand-selected. Participants were able to annotate images and provide comments. Supplemental materials include an image of the testing interface and precision, recall, and accuracy values for users. All participant results are in the codebase. Cohen's kappa was computed to assess inter-rater agreement between embryologists. This metric accounts for agreement occurring by chance, providing a more robust measure of consistency among raters.

**Classification Models**

To evaluate the clinical utility of synthetic images, we trained separate ResNet-50 classifiers for each Gardner category, as classification models often perform better when focused on a specific scoring task.

Baseline models trained exclusively on real images served as controls for three experiments: a class balance test, an augmentation test, and a partial replacement test. All experiments employed five-fold cross-validation using images from the 2022 WCM dataset. Real images were sourced from day 5 embryos, primarily captured between 108–112 HPI, with additional sampling up to 120 hours to increase the representation of rare scores. The dataset reflected FDs used in clinical practice: approximately 50% were acquired at FD 0, with 12.5% each at -30, -15, +15, and +30. Baseline dataset distributions are shown in Supplemental Figures 12, 13.

The class balance test began with a real-world distribution baseline and introduced synthetic images to underrepresented scores to achieve a balanced 4000-image training set as shown in Supplemental Figure 14. The augmentation test started from an evenly balanced 4000-image baseline, with synthetic images incrementally added in ~4000-image steps. The partial replacement test also began from the balanced baseline, replacing 20%, 40%, 60%, or 80% of real images with synthetic images matched for score and FD. Across all experiments, synthetic images were conditionally generated to match the FD distribution of the real image only datasets, and validation sets remained identical to their real-only baselines to ensure fair comparison.

All classification models were trained using ResNet-50 models with the Hugging Face Accelerate library. Models were initialized from ImageNet-pretrained ResNet-50 weights and fine-tuned using the AdamW optimizer, and a learning rate of .00001 with a 50 step constant-with-warmup scheduler. Training was conducted for 100 epochs with a batch size of 32. Cross-entropy loss was used for optimization, and model accuracy on the validation set was computed at each epoch.

### Computational Resources
We trained the VQ-VAE, LDMs, and downstream ResNet models each with one A6000 GPU.

### Data Availability
The final DIA models can be accessed at https://huggingface.co/ihlab/DIA for synthetic image generation.

### Code Availability
The code used in this study is available at https://github.com/naraharip2017/DIA/tree/main.

## Acknowledgements

We would like to acknowledge Laura Conversa, Laura Carrión, Savannah Phillips, and Shenni Liang for taking the Turing test.

## Author Contributions Statement

P.N., S.R., I.H. conceived the study. P.N. performed the experiments, method selection, analyses, and model development. P.N. wrote the code and computational analyses with input from S.R., I.H.. I.H., Q.Z., J.E.M., Z.R. and N.Z. provided the Weill Cornell datasets and labeled images. Q.Z., L.B., P.N., and

# Supplementary Material

| Score | Expansion Gardner Grade(s) | ICM Gardner Grade(s) | TE Gardner Grade(s) |
|---|---|---|---|
| 1 | 4 | A-, A | A-, A |
| 2 | 2-3, 3 | B-, B-/B, B | B-, B |
| 3 | 1-2, 2 | C, B-/C | C, B-/C |
| 4 | 1 | N/A | N/A |

**Supplemental Table 1:** Conversions between Gardner grades used at WCM and scores used during study

| | | | | | | | |
|---|---|---|---|---|---|---|---|
| | **Analysis: Real vs Synthetic Judgment** | | | | | | |
| | Class | precision | recall | support | | | |
| | Synthetic | 0.69 | 0.521 | 192 | | | |
| | Real | 0.615 | 0.766 | 192 | | | |
| | accuracy | | 0.643 | | | | |
| DIA-E | **Analysis: Expansion Scores \| Image Type: Real** | | | | **Analysis: Expansion Scores \| Image Type: Synthetic** | | |
| | Class | precision | recall | support | Class | precision | |
| | 1 | 0.638 | 0.625 | 48 | 1 | 0.774 | |
| | 2 | 0.372 | 0.333 | 48 | 2 | 0.675 | |
| | 3 | 0.346 | 0.375 | 48 | 3 | 0.561 | |
| | 4 | 0.52 | 0.542 | 48 | 4 | 0.714 | |
| | accuracy | | 0.469 | | accuracy | | 0.677 |
| | **Analysis: Real vs Synthetic Judgment** | | | | | | |
| | Class | precision | recall | support | | | |
| | Synthetic | 0.615 | 0.458 | 192 | | | |
| | Real | 0.568 | 0.714 | 192 | | | |
| | accuracy | | 0.586 | | | | |
| DIA-I | **Analysis: ICM Scores \| Image Type: Real** | | | | **Analysis: ICM Scores \| Image Type: Synthetic** | | |
| | Class | precision | recall | support | Class | precision | |
| | 1 | 0.805 | 0.516 | 64 | 1 | 0.925 | |
| | 2 | 0.483 | 0.438 | 64 | 2 | 0.636 | |
| | 3 | 0.581 | 0.844 | 64 | 3 | 0.643 | |
| | accuracy | | 0.599 | | accuracy | | 0.719 |
| DIA-T | **Analysis: Real vs Synthetic Judgment** | | | | | | |
| | Class | precision | recall | support | | | |
| | Synthetic | 0.619 | 0.474 | 192 | | | |

|  | Real | 0.574 | 0.708 | 192 | | | | |
|---|---|---|---|---|---|---|---|---|
|  | accuracy |  | 0.591 |  | | | | |
|  | **Analysis: TE Scores \| Image Type: Real** | | | | **Analysis: TE Scores \| Image Type: Synthetic** | | | |
|  | **Class** | **precision** | **recall** | **support** | **Class** | **precision** |  |  |
|  | 1 | 0.71 | 0.344 | 64 | 1 | 0.898 |  |  |
|  | 2 | 0.338 | 0.375 | 64 | 2 | 0.648 |  |  |
|  | 3 | 0.611 | 0.859 | 64 | 3 | 0.708 |  |  |
|  | accuracy |  | 0.526 |  | accuracy |  | 0.74 |  |
| **DIA-EIT** | **Analysis: Real vs Synthetic Judgment** | | | | | | | |
|  | **Class** | **precision** | **recall** | **support** | | | | |
|  | Synthetic | 0.672 | 0.455 | 288 | | | | |
|  | Real | 0.588 | 0.778 | 288 | | | | |
|  | accuracy |  | 0.616 |  | | | | |
|  | **Analysis: Expansion Scores \| Image Type: Real** | | | | **Analysis: Expansion Scores \| Image Type: Synthetic** | | | |
|  | **Class** | **precision** | **recall** | **support** | **Class** | **precision** |  |  |
|  | 1 | 0.562 | 0.75 | 72 | 1 | 0.607 |  |  |
|  | 2 | 0.263 | 0.347 | 72 | 2 | 0.34 |  |  |
|  | 3 | 0.231 | 0.208 | 72 | 3 | 0.217 |  |  |
|  | 4 | 0.75 | 0.333 | 72 | 4 | 0.939 |  |  |
|  | accuracy |  | 0.41 |  | accuracy |  | 0.493 |  |
|  | **Analysis: ICM Scores \| Image Type: Real** | | | | **Analysis: ICM Scores \| Image Type: Synthetic** | | | |
|  | **Class** | **precision** | **recall** | **support** | **Class** | **precision** |  |  |
|  | 1 | 0.687 | 0.479 | 96 | 1 | 0.815 |  |  |
|  | 2 | 0.482 | 0.552 | 96 | 2 | 0.463 |  |  |
|  | 3 | 0.631 | 0.729 | 96 | 3 | 0.602 |  |  |
|  | accuracy |  | 0.587 |  | accuracy |  | 0.604 |  |
|  | **Analysis: TE Scores \| Image Type: Real** | | | | **Analysis: TE Scores \| Image Type: Synthetic** | | | |
|  | **Class** | **precision** | **recall** | **support** | **Class** | **precision** |  |  |
|  | 1 | 0.78 | 0.333 | 96 | 1 | 0.889 |  |  |
|  | 2 | 0.442 | 0.552 | 96 | 2 | 0.57 |  |  |
|  | 3 | 0.669 | 0.885 | 96 | 3 | 0.758 |  |  |
|  | accuracy |  | 0.59 |  | accuracy |  | 0.708 |  |

**Supplemental Table 2:** Precision, Recall, and Accuracy values for Turing Test conducted by embryologists. Scoring metrics are separated by ground-truth real and synthetic images.

| MODEL | True Images: Real | | | | | True Images: Synthetic | | | | |
|---|---|---|---|---|---|---|---|---|---|---|
| DIA-E | Agreement on: Real vs. Synthetic Judgment | | | | | | | | | |
| | Embryologist | 1 | 2 | 3 | 4 | Embryologist | 1 | 2 | 3 | 4 |
| | 1 | 1 | 0 | 0.3438 | -0.0364 | 1 | 1 | 0.1564 | 0.2 | 0.1399 |
| | 2 | 0 | 1 | 0.1111 | -0.1556 | 2 | 0.1564 | 1 | 0.0562 | 0.0712 |
| | 3 | 0.3438 | 0.1111 | 1 | 0.0727 | 3 | 0.2 | 0.0562 | 1 | 0.0943 |
| | 4 | -0.0364 | -0.1556 | 0.0727 | 1 | 4 | 0.1399 | 0.0712 | 0.0943 | 1 |
| | Agreement on: Expansion Score | | | | | | | | | |
| | Embryologist | 1 | 2 | 3 | 4 | Embryologist | 1 | 2 | 3 | 4 |
| | 1 | 1 | 0.2645 | 0.2049 | 0.1978 | 1 | 1 | 0.4762 | 0.5032 | 0.3259 |
| | 2 | 0.2645 | 1 | 0.5519 | 0.4425 | 2 | 0.4762 | 1 | 0.7445 | 0.5545 |
| | 3 | 0.2049 | 0.5519 | 1 | 0.6052 | 3 | 0.5032 | 0.7445 | 1 | 0.5558 |
| | 4 | 0.1978 | 0.4425 | 0.6052 | 1 | 4 | 0.3259 | 0.5545 | 0.5558 | 1 |
| DIA-I | Agreement on: Real vs. Synthetic Judgment | | | | | | | | | |
| | Embryologist | 1 | 2 | 3 | 4 | Embryologist | 1 | 2 | 3 | 4 |
| | 1 | 1 | 0.0417 | 0.0833 | -0.0833 | 1 | 1 | -0.074 | 0.0316 | 0.058 |
| | 2 | 0.0417 | 1 | -0.0857 | -0.1087 | 2 | -0.074 | 1 | 0.0891 | 0.4068 |
| | 3 | 0.0833 | -0.0857 | 1 | 0.125 | 3 | 0.0316 | 0.0891 | 1 | 0.1698 |
| | 4 | -0.0833 | -0.1087 | 0.125 | 1 | 4 | 0.058 | 0.4068 | 0.1698 | 1 |
| | Agreement on: ICM Score | | | | | | | | | |
| | Embryologist | 1 | 2 | 3 | 4 | Embryologist | 1 | 2 | 3 | 4 |
| | 1 | 1 | 0.4188 | 0.5059 | 0.265 | 1 | 1 | 0.5251 | 0.5755 | 0.5112 |
| | 2 | 0.4188 | 1 | 0.6333 | 0.2441 | 2 | 0.5251 | 1 | 0.6549 | 0.6286 |
| | 3 | 0.5059 | 0.6333 | 1 | 0.3684 | 3 | 0.5755 | 0.6549 | 1 | 0.6508 |
| | 4 | 0.265 | 0.2441 | 0.3684 | 1 | 4 | 0.5112 | 0.6286 | 0.6508 | 1 |
| DIA-T | Agreement on: Real vs. Synthetic Judgment | | | | | | | | | |
| | Embryologist | 1 | 2 | 3 | 4 | Embryologist | 1 | 2 | 3 | 4 |
| | 1 | 1 | 0.1851 | 0.1738 | 0.0384 | 1 | 1 | 0.2394 | 0.1064 | 0.0886 |
| | 2 | 0.1851 | 1 | 0.0607 | -0.0397 | 2 | 0.2394 | 1 | 0.1158 | 0.3642 |
| | 3 | 0.1738 | 0.0607 | 1 | 0.0629 | 3 | 0.1064 | 0.1158 | 1 | -0.0485 |
| | 4 | 0.0384 | -0.0397 | 0.0629 | 1 | 4 | 0.0886 | 0.3642 | -0.0485 | 1 |
| | Agreement on: TE Score | | | | | | | | | |
| | Embryologist | 1 | 2 | 3 | 4 | Embryologist | 1 | 2 | 3 | 4 |
| | 1 | 1 | 0.5702 | 0.4696 | 0.4523 | 1 | 1 | 0.6701 | 0.6113 | 0.3156 |
| | 2 | 0.5702 | 1 | 0.6883 | 0.6537 | 2 | 0.6701 | 1 | 0.776 | 0.4818 |
| | 3 | 0.4696 | 0.6883 | 1 | 0.5633 | 3 | 0.6113 | 0.776 | 1 | 0.456 |
| | 4 | 0.4523 | 0.6537 | 0.5633 | 1 | 4 | 0.3156 | 0.4818 | 0.456 | 1 |
| DIA-EIT | Agreement on: Real vs. Synthetic Judgment | | | | | | | | | |

| Embryologist | 1 | 2 | 3 | 4 | Embryologist | 1 | 2 | 3 | 4 |
|---|---|---|---|---|---|---|---|---|---|
| 1 | 1 | 0.1525 | 0.1181 | 0.069 | 1 | 1 | 0.2778 | 0.1455 | -0.0253 |
| 2 | 0.1525 | 1 | 0.1148 | 0.3662 | 2 | 0.2778 | 1 | 0.0833 | 0.3333 |
| 3 | 0.1181 | 0.1148 | 1 | 0.078 | 3 | 0.1455 | 0.0833 | 1 | 0.032 |
| 4 | 0.069 | 0.3662 | 0.078 | 1 | 4 | -0.0253 | 0.3333 | 0.032 | 1 |
| Agreement on: Expansion Score | | | | | | | | | |
| Embryologist | 1 | 2 | 3 | 4 | Embryologist | 1 | 2 | 3 | 4 |
| 1 | 1 | 0.4833 | 0.516 | 0.4297 | 1 | 1 | 0.6988 | 0.4922 | 0.5663 |
| 2 | 0.4833 | 1 | 0.5033 | 0.5599 | 2 | 0.6988 | 1 | 0.6771 | 0.6112 |
| 3 | 0.516 | 0.5033 | 1 | 0.553 | 3 | 0.4922 | 0.6771 | 1 | 0.454 |
| 4 | 0.4297 | 0.5599 | 0.553 | 1 | 4 | 0.5663 | 0.6112 | 0.454 | 1 |
| Agreement on: ICM Score | | | | | | | | | |
| Embryologist | 1 | 2 | 3 | 4 | Embryologist | 1 | 2 | 3 | 4 |
| 1 | 1 | 0.4735 | 0.3895 | 0.3668 | 1 | 1 | 0.5596 | 0.4434 | 0.4306 |
| 2 | 0.4735 | 1 | 0.4244 | 0.2083 | 2 | 0.5596 | 1 | 0.6181 | 0.4352 |
| 3 | 0.3895 | 0.4244 | 1 | 0.3149 | 3 | 0.4434 | 0.6181 | 1 | 0.4997 |
| 4 | 0.3668 | 0.2083 | 0.3149 | 1 | 4 | 0.4306 | 0.4352 | 0.4997 | 1 |
| Agreement on: TE Score | | | | | | | | | |
| Embryologist | 1 | 2 | 3 | 4 | Embryologist | 1 | 2 | 3 | 4 |
| 1 | 1 | 0.5872 | 0.5096 | 0.5047 | 1 | 1 | 0.745 | 0.6253 | 0.3157 |
| 2 | 0.5872 | 1 | 0.5779 | 0.4581 | 2 | 0.745 | 1 | 0.7887 | 0.4611 |
| 3 | 0.5096 | 0.5779 | 1 | 0.4182 | 3 | 0.6253 | 0.7887 | 1 | 0.5996 |
| 4 | 0.5047 | 0.4581 | 0.4182 | 1 | 4 | 0.3157 | 0.4611 | 0.5996 | 1 |

**Supplemental Table 3:** Cohen's Kappa agreement scores for embryologists for real vs synthetic judgement and scores for each model. Metrics are separated by ground-truth real and synthetic images.

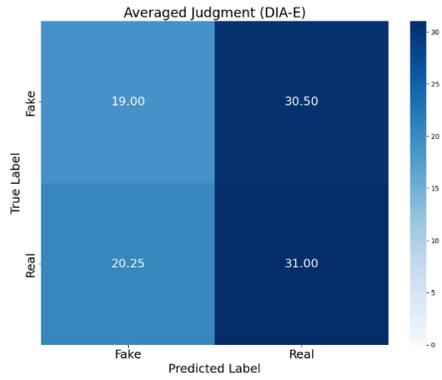
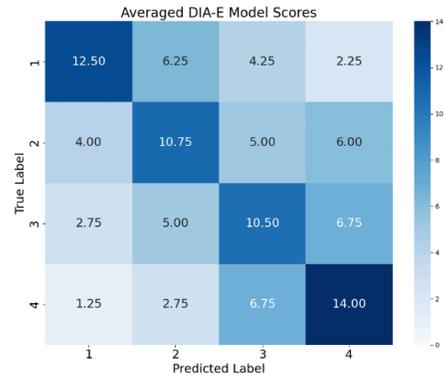

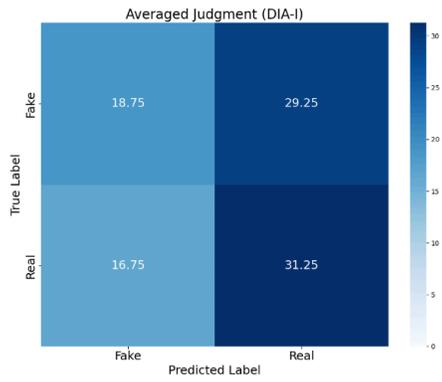
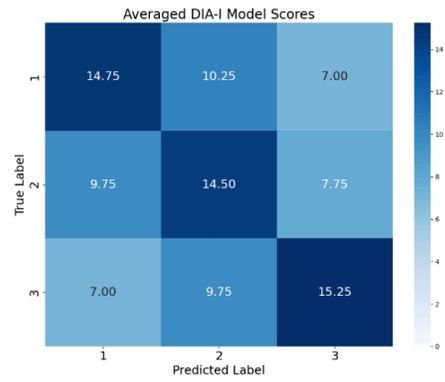

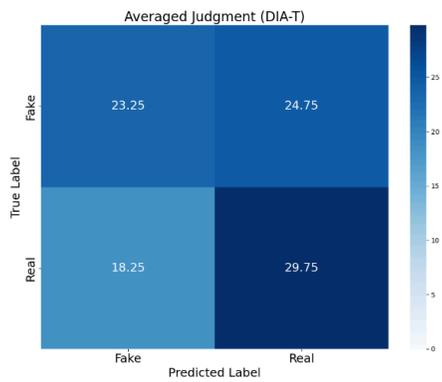
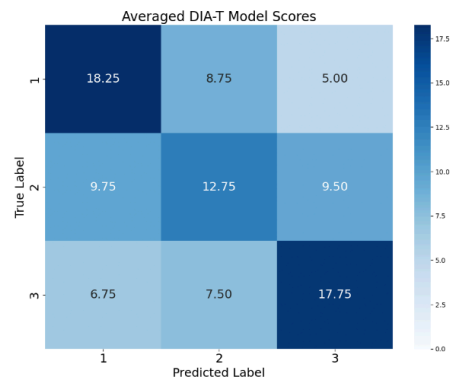

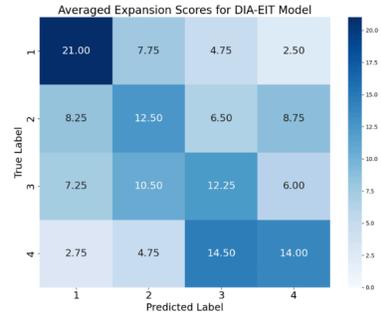
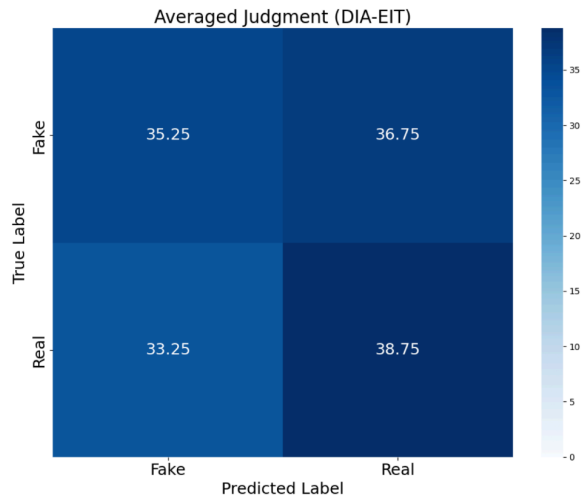
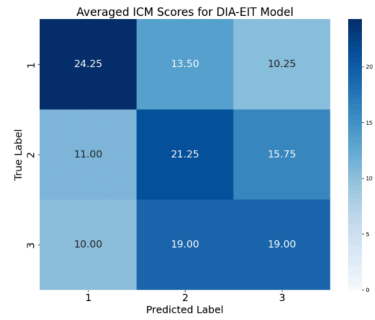
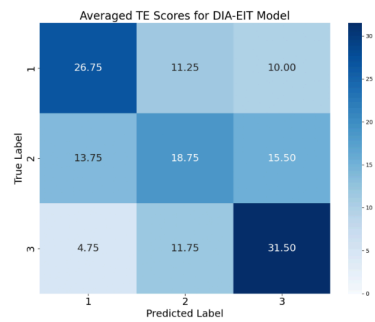

**Supplemental Figure 1:** Confusion matrices from the Turing test averaged across four non-embryologists. The matrices on the left show the classification of real versus synthetic images. The matrices on the right show predictions of embryo quality scores compared to ground truth for the relevant scoring categories of each model.

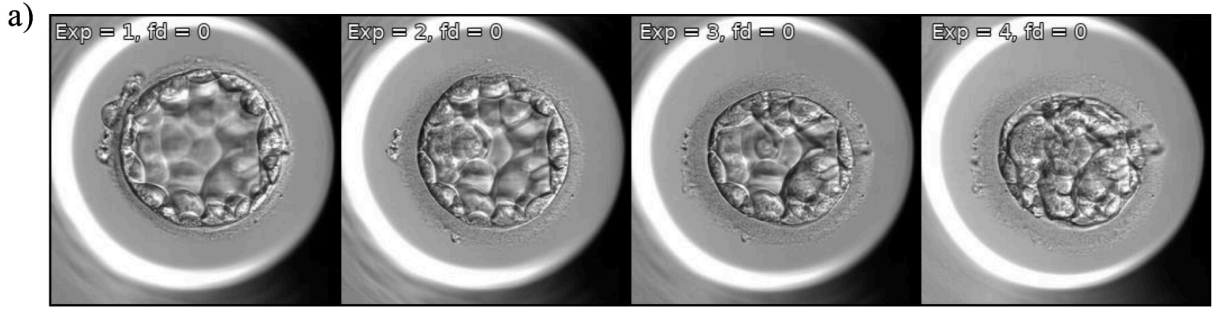

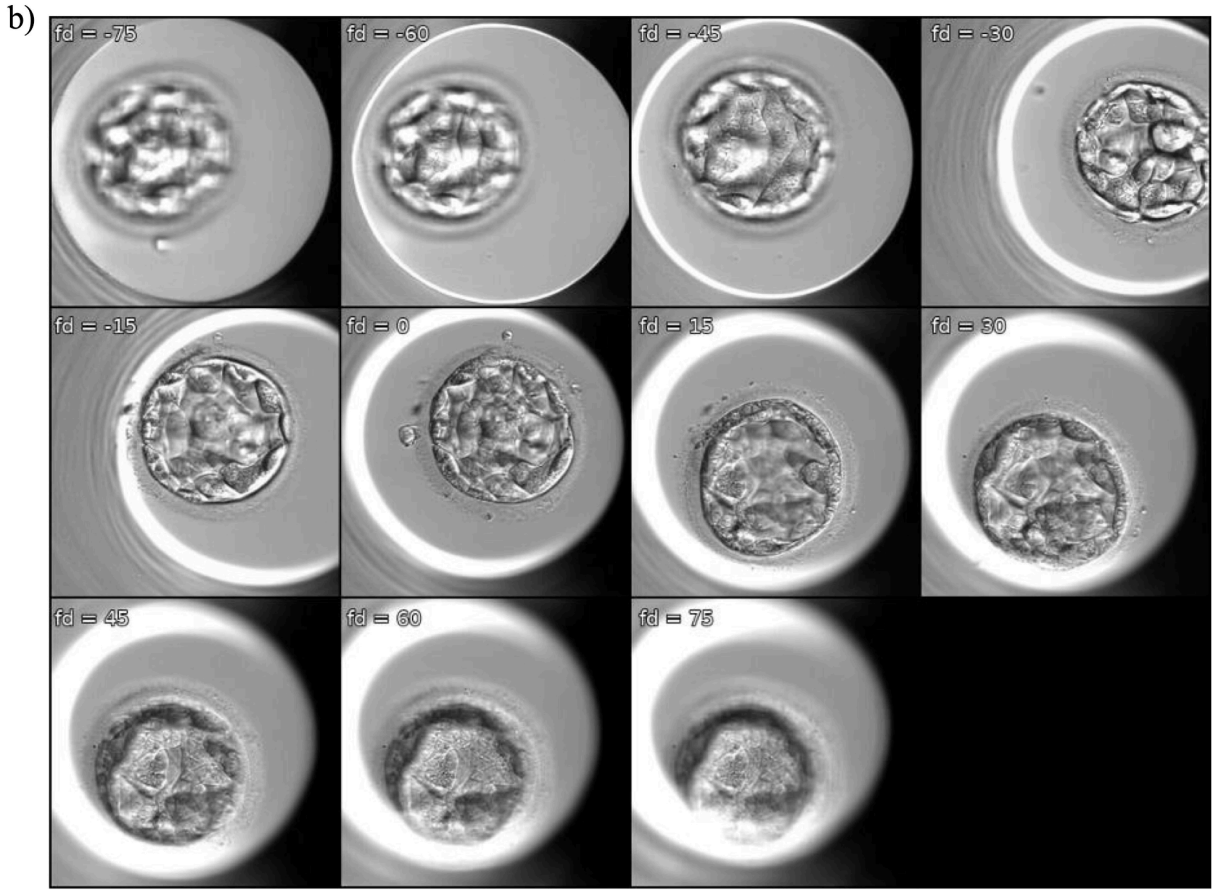

**Supplemental Figure 2: a)** Fixed Gaussian noise sample passed through DIA-E while varying only expansion score **b)** Fixed Gaussian noise sample passed through DIA-E while varying just z-axis focal depth.

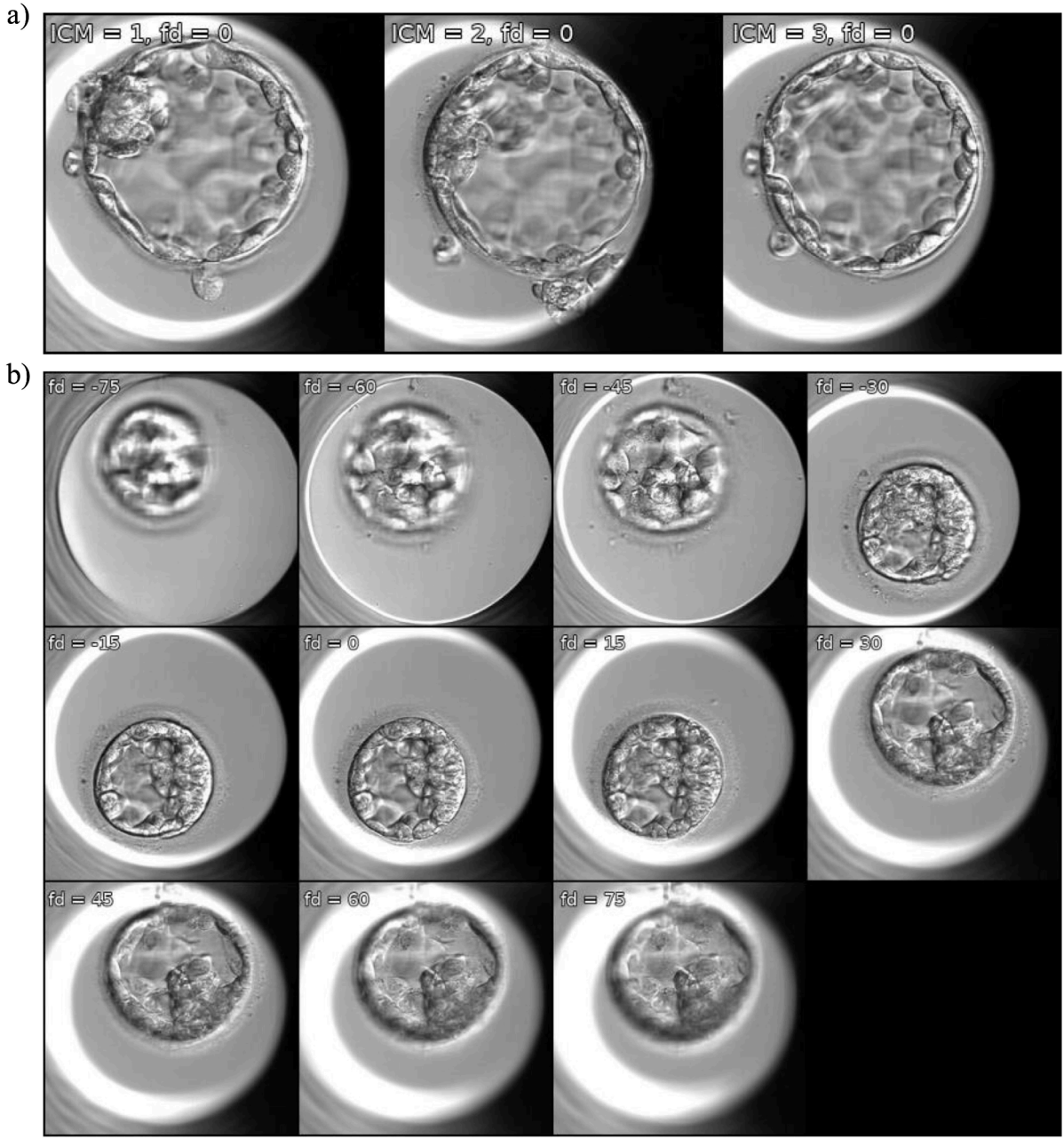

**Supplemental Figure 3: a)** Fixed Gaussian noise sample passed through DIA-I while varying only ICM score **b)** Fixed Gaussian noise sample passed through DIA-I while varying only z-axis focal depth.

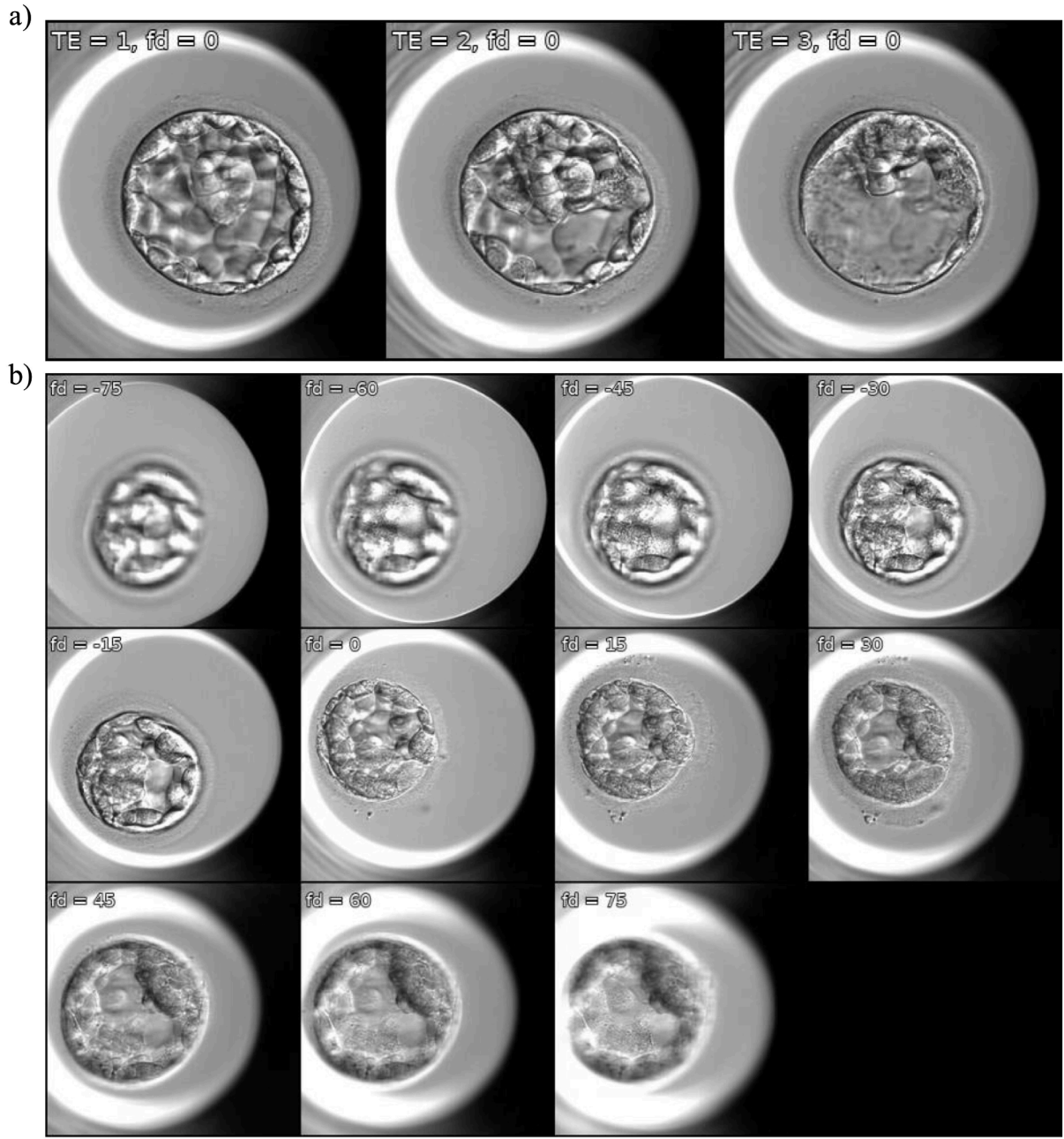

**Supplemental Figure 4: a)** Fixed Gaussian noise sample passed through DIA-T while varying only ICM score **b)** Fixed Gaussian noise sample passed through DIA-T while varying only z-axis focal depth.

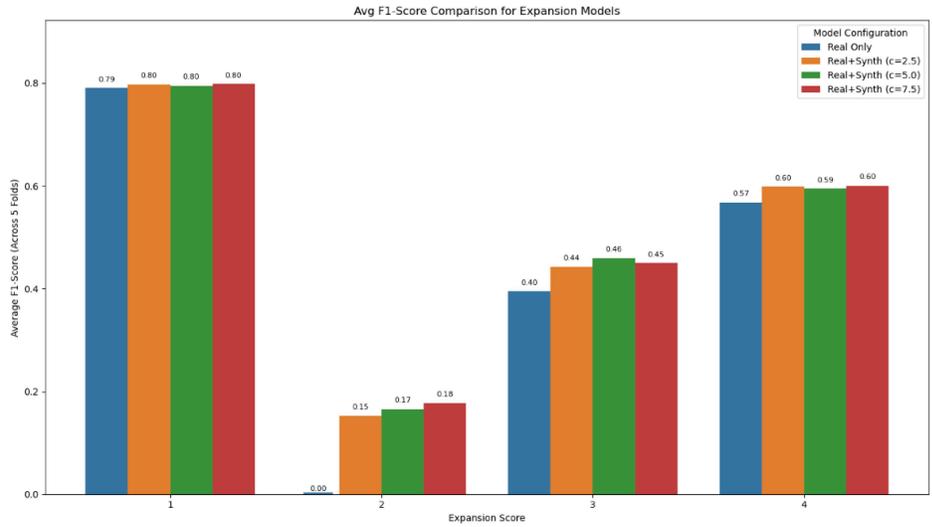
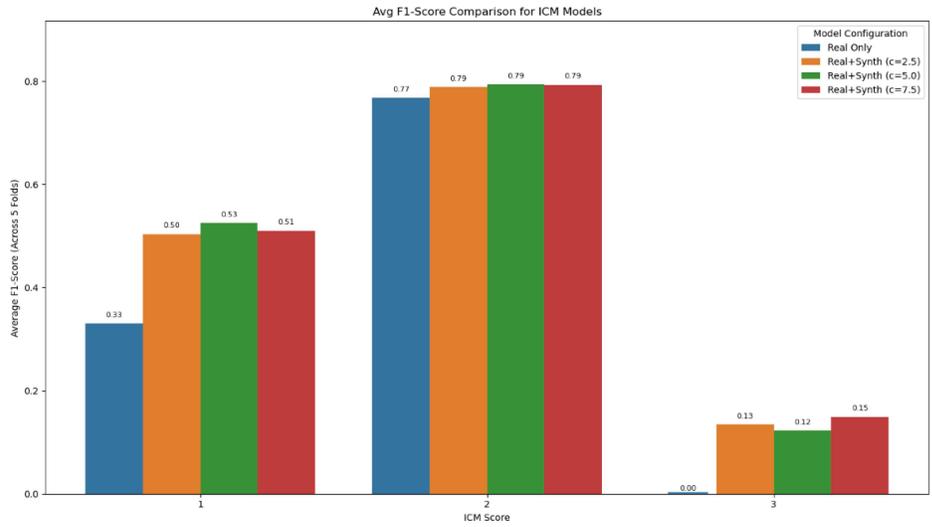
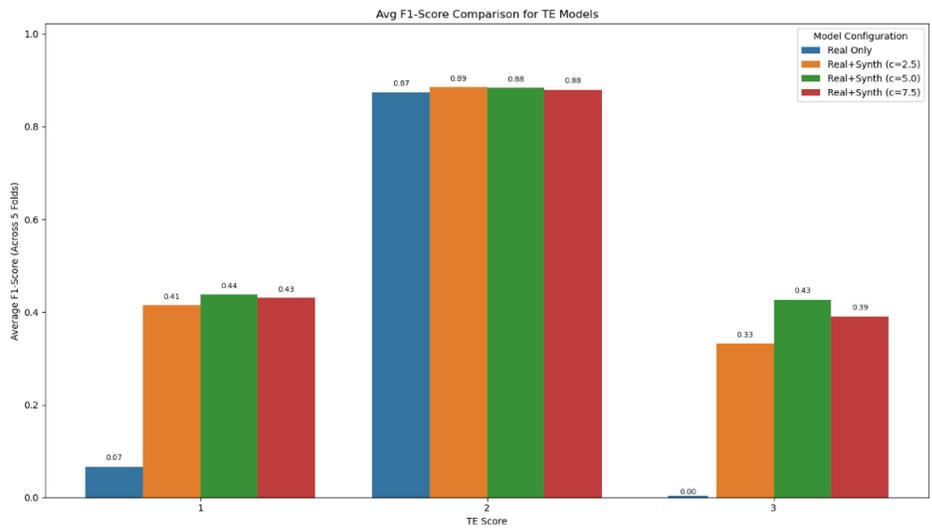

**Supplemental Figure 5:** F1 score comparison for class balance test for each model

| Scoring Category | Images | CFG-Scale | Fold 1 Acc | Fold 2 Acc | Fold 3 Acc | Fold 4 Acc | Fold 5 Acc | p-value | t-stat |
|---|---|---|---|---|---|---|---|---|---|
| **Class Balance Test** | | | | | | | | | |
| Expansion | Real Baseline | - | 0.6033 | 0.6166 | 0.6200 | 0.6057 | 0.6227 | - | - |
| | Baseline + Synthetic | 2.5000 | 0.6284 | 0.6247 | 0.6460 | 0.6324 | 0.6430 | 0.0036 | 6.1263 |
| | | 5.0000 | 0.6117 | 0.6349 | 0.6700 | 0.6345 | 0.6207 | 0.0819 | 2.3116 |
| | | 7.5000 | 0.6409 | 0.6430 | 0.6440 | 0.6386 | 0.6227 | 0.0205 | 3.7186 |
| ICM | Real Baseline | - | 0.7775 | 0.7778 | 0.7788 | 0.7723 | 0.7708 | - | - |
| | Baseline + Synthetic | 2.5000 | 0.7897 | 0.8140 | 0.8029 | 0.8119 | 0.7985 | 0.0044 | 5.7862 |
| | | 5.0000 | 0.7995 | 0.7899 | 0.8221 | 0.8119 | 0.7909 | 0.0102 | 4.5791 |
| | | 7.5000 | 0.7946 | 0.8237 | 0.7981 | 0.8094 | 0.8111 | 0.0053 | 5.5056 |
| TE | Real Baseline | - | 0.6033 | 0.6166 | 0.6200 | 0.6057 | 0.6227 | - | - |
| | Baseline + Synthetic | 2.5000 | 0.6653 | 0.6633 | 0.6800 | 0.6667 | 0.7176 | 0.0034 | 6.2099 |
| | | 5.0000 | 0.6633 | 0.6673 | 0.7032 | 0.6787 | 0.7353 | 0.0077 | 4.9579 |
| | | 7.5000 | 0.6613 | 0.6574 | 0.6821 | 0.6767 | 0.7137 | 0.0076 | 4.9706 |
| **Augmentation Test** | | | | | | | | | |
| Expansion | Real Baseline | - | 0.6066 | 0.5804 | 0.6086 | 0.6143 | 0.6092 | - | - |
| | Baseline + 4000 Synthetic Images | 2.5000 | 0.6167 | 0.6258 | 0.6268 | 0.6636 | 0.6455 | 0.0141 | 4.1601 |
| | | 5.0000 | 0.6328 | 0.6298 | 0.6338 | 0.6586 | 0.6425 | 0.0018 | 7.3702 |
| | | 7.5000 | 0.6439 | 0.6258 | 0.6388 | 0.6546 | 0.6304 | 0.0012 | 8.2850 |
| | Baseline + 8000 Synthetic Images | 2.5000 | 0.6388 | 0.6237 | 0.6489 | 0.6596 | 0.6596 | 0.0001 | 14.0423 |
| | | 5.0000 | 0.6730 | 0.6479 | 0.6831 | 0.6818 | 0.6647 | 0.0000 | 21.5981 |
| | | 7.5000 | 0.6861 | 0.6358 | 0.6630 | 0.7090 | 0.6788 | 0.0007 | 9.2897 |
| | Baseline + 12000 Synthetic Images | 2.5000 | 0.6771 | 0.6590 | 0.6881 | 0.6677 | 0.6878 | 0.0001 | 14.5567 |
| | | 5.0000 | 0.7123 | 0.6831 | 0.6962 | 0.7009 | 0.7080 | 0.0000 | 24.6925 |
| | | 7.5000 | 0.6891 | 0.6811 | 0.6932 | 0.7110 | 0.7009 | 0.0000 | 26.3673 |
| | Baseline + 16000 Synthetic Images | 2.5000 | 0.7052 | 0.6922 | 0.7022 | 0.7009 | 0.6838 | 0.0001 | 15.0605 |
| | | 5.0000 | 0.7123 | 0.6911 | 0.6972 | 0.7090 | 0.6989 | 0.0000 | 22.1700 |
| | | 7.5000 | 0.7193 | 0.6811 | 0.7233 | 0.7039 | 0.6949 | 0.0001 | 17.1374 |
| | Baseline + 20000 Synthetic Images | 2.5000 | 0.6740 | 0.6932 | 0.7193 | 0.7009 | 0.7251 | 0.0002 | 13.9304 |
| | | 5.0000 | 0.7183 | 0.6811 | 0.7304 | 0.7190 | 0.6989 | 0.0008 | 9.0300 |
| | | 7.5000 | 0.7314 | 0.7012 | 0.7314 | 0.7341 | 0.6908 | 0.0010 | 8.6564 |
| ICM | Real Baseline | - | 0.6509 | 0.6639 | 0.6730 | 0.6475 | 0.6837 | - | - |
| | Baseline + 4000 Synthetic Images | 2.5000 | 0.7193 | 0.7042 | 0.6962 | 0.6999 | 0.7261 | 0.0037 | 6.0801 |
| | | 5.0000 | 0.7143 | 0.7213 | 0.7294 | 0.7221 | 0.7382 | 0.0001 | 16.7407 |
| | | 7.5000 | 0.7042 | 0.7113 | 0.7193 | 0.7200 | 0.7251 | 0.0007 | 9.5833 |
| | Baseline + 8000 Synthetic Images | 2.5000 | 0.7243 | 0.7143 | 0.7304 | 0.7362 | 0.7704 | 0.0007 | 9.2991 |
| | | 5.0000 | 0.7404 | 0.7183 | 0.7425 | 0.7412 | 0.7291 | 0.0017 | 7.4354 |
| | | 7.5000 | 0.7455 | 0.7163 | 0.7394 | 0.7402 | 0.7462 | 0.0010 | 8.7066 |
| | Baseline + 12000 Synthetic Images | 2.5000 | 0.7505 | 0.7203 | 0.7495 | 0.7301 | 0.7452 | 0.0006 | 9.7324 |
| | | 5.0000 | 0.7485 | 0.7404 | 0.7565 | 0.7543 | 0.7513 | 0.0003 | 12.1841 |
| | | 7.5000 | 0.7344 | 0.7364 | 0.7445 | 0.7402 | 0.7674 | 0.0000 | 20.3974 |
| | Baseline + 16000 Synthetic Images | 2.5000 | 0.7243 | 0.7384 | 0.7495 | 0.7462 | 0.7744 | 0.0001 | 16.3604 |
| | | 5.0000 | 0.7596 | 0.7344 | 0.7555 | 0.7603 | 0.7603 | 0.0005 | 10.4549 |
| | | 7.5000 | 0.7445 | 0.7243 | 0.7616 | 0.7644 | 0.7754 | 0.0006 | 10.0359 |
| | Baseline + 20000 Synthetic Images | 2.5000 | 0.7475 | 0.7354 | 0.7646 | 0.7613 | 0.7795 | 0.0001 | 17.0469 |
| | | 5.0000 | 0.7334 | 0.7394 | 0.7817 | 0.7694 | 0.7664 | 0.0004 | 11.1905 |
| | | 7.5000 | 0.7616 | 0.7344 | 0.7535 | 0.7644 | 0.7784 | 0.0001 | 15.6304 |
| | Baseline + 24000 Synthetic Images | 2.5000 | 0.7656 | 0.7344 | 0.7746 | 0.7482 | 0.8056 | 0.0003 | 11.5552 |
| | | 5.0000 | 0.7525 | 0.7505 | 0.7857 | 0.7633 | 0.8026 | 0.0001 | 18.1232 |
| | | 7.5000 | 0.7777 | 0.7435 | 0.7757 | 0.7825 | 0.7724 | 0.0006 | 9.9630 |
| TE | Real Baseline | - | 0.7313 | 0.7505 | 0.7645 | 0.7542 | 0.7552 | - | - |
| | Baseline + 4000 Synthetic Images | 2.5000 | 0.7827 | 0.7777 | 0.8028 | 0.7835 | 0.7764 | 0.0031 | 6.3675 |
| | | 5.0000 | 0.7757 | 0.7666 | 0.7998 | 0.7885 | 0.7875 | 0.0021 | 7.0896 |
| | | 7.5000 | 0.7676 | 0.7746 | 0.8068 | 0.7633 | 0.7754 | 0.0110 | 4.4803 |
| | Baseline + 8000 Synthetic Images | 2.5000 | 0.7837 | 0.7897 | 0.8038 | 0.7784 | 0.7976 | 0.0009 | 8.7432 |
| | | 5.0000 | 0.7857 | 0.7998 | 0.7998 | 0.7905 | 0.7895 | 0.0005 | 10.0888 |
| | | 7.5000 | 0.7928 | 0.8048 | 0.7948 | 0.7704 | 0.7654 | 0.0277 | 3.3837 |
| | Baseline + 12000 Synthetic Images | 2.5000 | 0.8028 | 0.7968 | 0.8149 | 0.8147 | 0.7885 | 0.0013 | 8.0958 |
| | | 5.0000 | 0.8028 | 0.7918 | 0.8249 | 0.8036 | 0.8066 | 0.0004 | 10.6084 |
| | | 7.5000 | 0.7787 | 0.7918 | 0.8058 | 0.8056 | 0.7996 | 0.0000 | 23.4004 |
| | Baseline + 16000 Synthetic Images | 2.5000 | 0.8109 | 0.8048 | 0.8199 | 0.8046 | 0.8137 | 0.0003 | 11.5950 |
| | | 5.0000 | 0.8109 | 0.8129 | 0.8169 | 0.8026 | 0.8187 | 0.0003 | 11.3155 |
| | | 7.5000 | 0.8139 | 0.8018 | 0.8169 | 0.8006 | 0.8137 | 0.0008 | 9.1210 |
| | Baseline + 20000 Synthetic Images | 2.5000 | 0.8139 | 0.8139 | 0.8280 | 0.8036 | 0.8248 | 0.0007 | 9.5905 |
| | | 5.0000 | 0.8058 | 0.8169 | 0.8169 | 0.8107 | 0.8207 | 0.0003 | 11.3090 |
| | | 7.5000 | 0.8129 | 0.8209 | 0.8330 | 0.8157 | 0.8369 | 0.0008 | 9.0923 |
| **Partial Replace Test** | | | | | | | | | | |
| Expansion | Real Baseline | - | 0.6066 | 0.5805 | 0.6087 | 0.6143 | 0.6093 | - | - |
| | 80% Baseline 20% Synthetic Images | 2.5000 | 0.5855 | 0.5654 | 0.5845 | 0.5801 | 0.5801 | 0.0017 | -7.5225 |
| | | 5.0000 | 0.5956 | 0.5775 | 0.5966 | 0.6093 | 0.5932 | 0.0168 | -3.9500 |
| | | 7.5000 | 0.5805 | 0.5674 | 0.5885 | 0.5861 | 0.6133 | 0.0453 | -2.8731 |
| | 60% Baseline 40% Synthetic Images | 2.5000 | 0.5855 | 0.5523 | 0.5915 | 0.5942 | 0.5660 | 0.0052 | -5.5281 |
| | | 5.0000 | 0.5825 | 0.5523 | 0.5835 | 0.6083 | 0.5700 | 0.0101 | -4.5879 |
| | | 7.5000 | 0.5775 | 0.5734 | 0.5775 | 0.6042 | 0.5478 | 0.0460 | -2.8581 |
| | 40% Baseline 60% Synthetic Images | 2.5000 | 0.5624 | 0.5252 | 0.5624 | 0.5861 | 0.5579 | 0.0006 | -9.7049 |
| | | 5.0000 | 0.5654 | 0.5412 | 0.5624 | 0.5740 | 0.5488 | 0.0003 | -11.5954 |
| | | 7.5000 | 0.5674 | 0.5654 | 0.5634 | 0.5831 | 0.5670 | 0.0031 | -6.3927 |
| | 20% Baseline 80% Synthetic Images | 2.5000 | 0.5423 | 0.5151 | 0.5453 | 0.5227 | 0.5267 | 0.0002 | -12.7642 |
| | | 5.0000 | 0.5412 | 0.5463 | 0.5433 | 0.5599 | 0.5196 | 0.0023 | -6.8736 |
| | | 7.5000 | 0.5392 | 0.5382 | 0.5362 | 0.5217 | 0.5418 | 0.0010 | -8.5229 |
| ICM | Real Baseline | - | 0.6509 | 0.6640 | 0.6730 | 0.6475 | 0.6838 | - | - |
| | 80% Baseline 20% Synthetic Images | 2.5000 | 0.6378 | 0.6298 | 0.6831 | 0.6465 | 0.6928 | 0.5177 | -0.7086 |
| | | 5.0000 | 0.6479 | 0.6439 | 0.6911 | 0.6596 | 0.6828 | 0.8643 | 0.1822 |
| | | 7.5000 | 0.6358 | 0.6348 | 0.6650 | 0.6757 | 0.6707 | 0.4804 | -0.7773 |
| | 60% Baseline 40% Synthetic Images | 2.5000 | 0.6449 | 0.6408 | 0.6660 | 0.6606 | 0.6999 | 0.8554 | -0.1943 |
| | | 5.0000 | 0.6348 | 0.6398 | 0.6740 | 0.6395 | 0.6697 | 0.0431 | -2.9225 |
| | | 7.5000 | 0.6227 | 0.6378 | 0.6469 | 0.6596 | 0.6999 | 0.3569 | -1.0403 |
| | 40% Baseline 60% Synthetic Images | 2.5000 | 0.6076 | 0.6097 | 0.6097 | 0.6183 | 0.6324 | 0.0011 | -8.3891 |
| | | 5.0000 | 0.6398 | 0.6237 | 0.6569 | 0.6173 | 0.6546 | 0.0084 | -4.8372 |
| | | 7.5000 | 0.6177 | 0.6217 | 0.6187 | 0.6123 | 0.6445 | 0.0004 | -10.9995 |
| | 20% Baseline 80% Synthetic Images | 2.5000 | 0.5966 | 0.5855 | 0.5714 | 0.5831 | 0.6244 | 0.0011 | -8.4279 |
| | | 5.0000 | 0.6197 | 0.6107 | 0.6328 | 0.5801 | 0.6354 | 0.0014 | -7.8546 |
| | | 7.5000 | 0.5966 | 0.6127 | 0.6006 | 0.5811 | 0.6133 | 0.0001 | -14.6660 |
| TE | Real Baseline | - | 0.7314 | 0.7505 | 0.7646 | 0.7543 | 0.7553 | - | - |
| | 80% Baseline 20% Synthetic Images | 2.5000 | 0.7414 | 0.7475 | 0.7465 | 0.7382 | 0.7543 | 0.3390 | -1.0849 |
| | | 5.0000 | 0.7133 | 0.7535 | 0.7435 | 0.7422 | 0.7341 | 0.0376 | -3.0604 |
| | | 7.5000 | 0.7324 | 0.7535 | 0.7636 | 0.7432 | 0.7553 | 0.5478 | -0.6557 |
| | 60% Baseline 40% Synthetic Images | 2.5000 | 0.7183 | 0.7264 | 0.7414 | 0.7291 | 0.7150 | 0.0045 | -5.7749 |
| | | 5.0000 | 0.7445 | 0.7213 | 0.7334 | 0.7241 | 0.7382 | 0.0873 | -2.2535 |
| | | 7.5000 | 0.7082 | 0.7314 | 0.7485 | 0.7090 | 0.7331 | 0.0083 | -4.8514 |
| | 40% Baseline 60% Synthetic Images | 2.5000 | 0.7254 | 0.7082 | 0.7183 | 0.6999 | 0.7241 | 0.0126 | -4.3018 |
| | | 5.0000 | 0.7123 | 0.7163 | 0.7213 | 0.6989 | 0.7180 | 0.0031 | -6.3903 |
| | | 7.5000 | 0.6791 | 0.7183 | 0.7344 | 0.7019 | 0.7251 | 0.0017 | -7.4811 |
| | 20% Baseline 80% Synthetic Images | 2.5000 | 0.6901 | 0.7042 | 0.7103 | 0.7080 | 0.7029 | 0.0000 | -20.4766 |
| | | 5.0000 | 0.6901 | 0.6972 | 0.7022 | 0.6838 | 0.7120 | 0.0006 | -9.7286 |
| | | 7.5000 | 0.6942 | 0.6972 | 0.7153 | 0.6737 | 0.7090 | 0.0019 | -7.3053 |

**Supplemental Table 4:** Individual fold accuracies for ResNet experiments for Class Balance, Augmentation, and Partial Replacement tests.

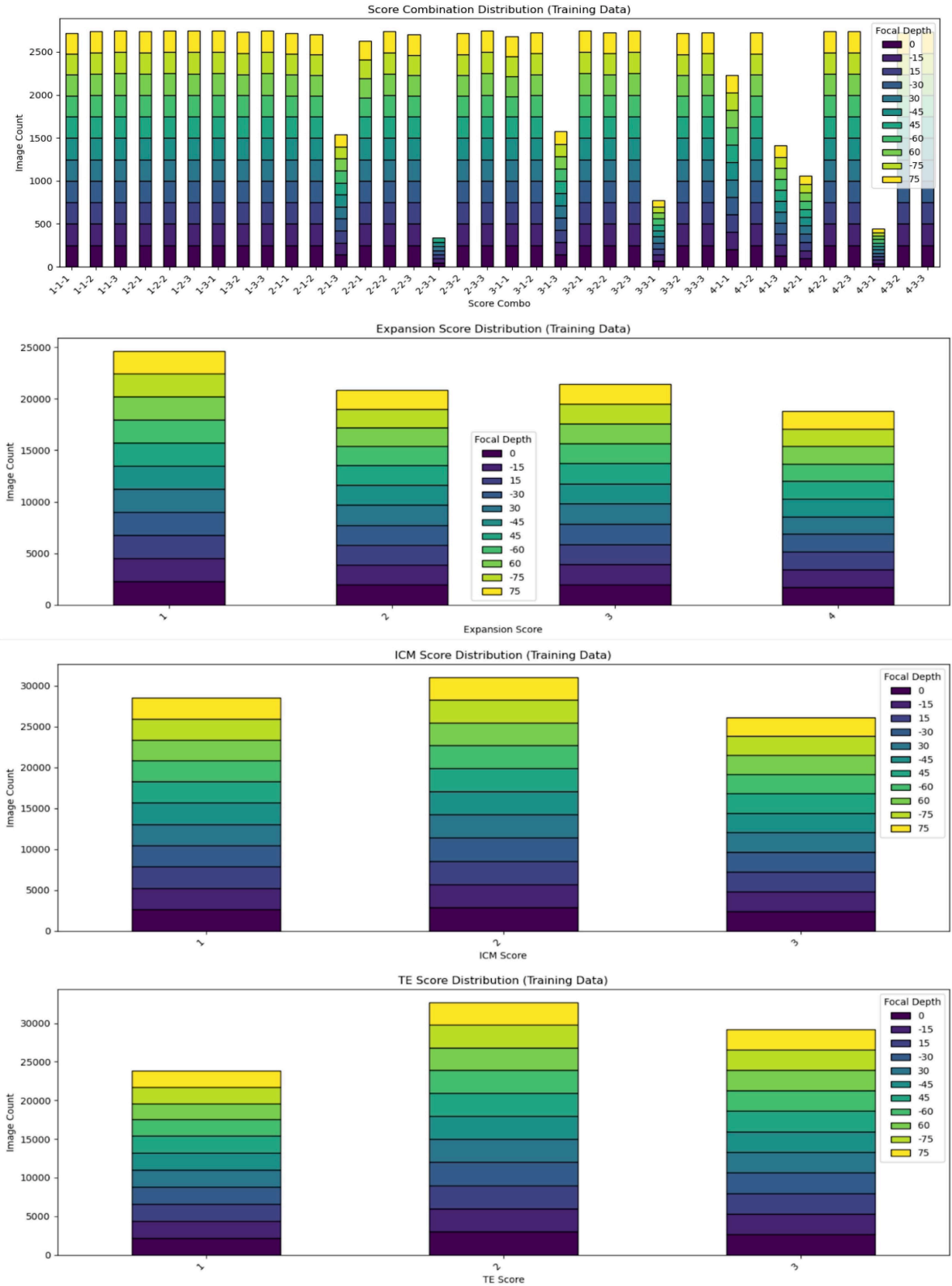

**Supplemental Figure 6:** DIA training data distribution across cumulative scores and individual scoring categories. Score combo in the first graph refers to (Expansion-ICM-TE)

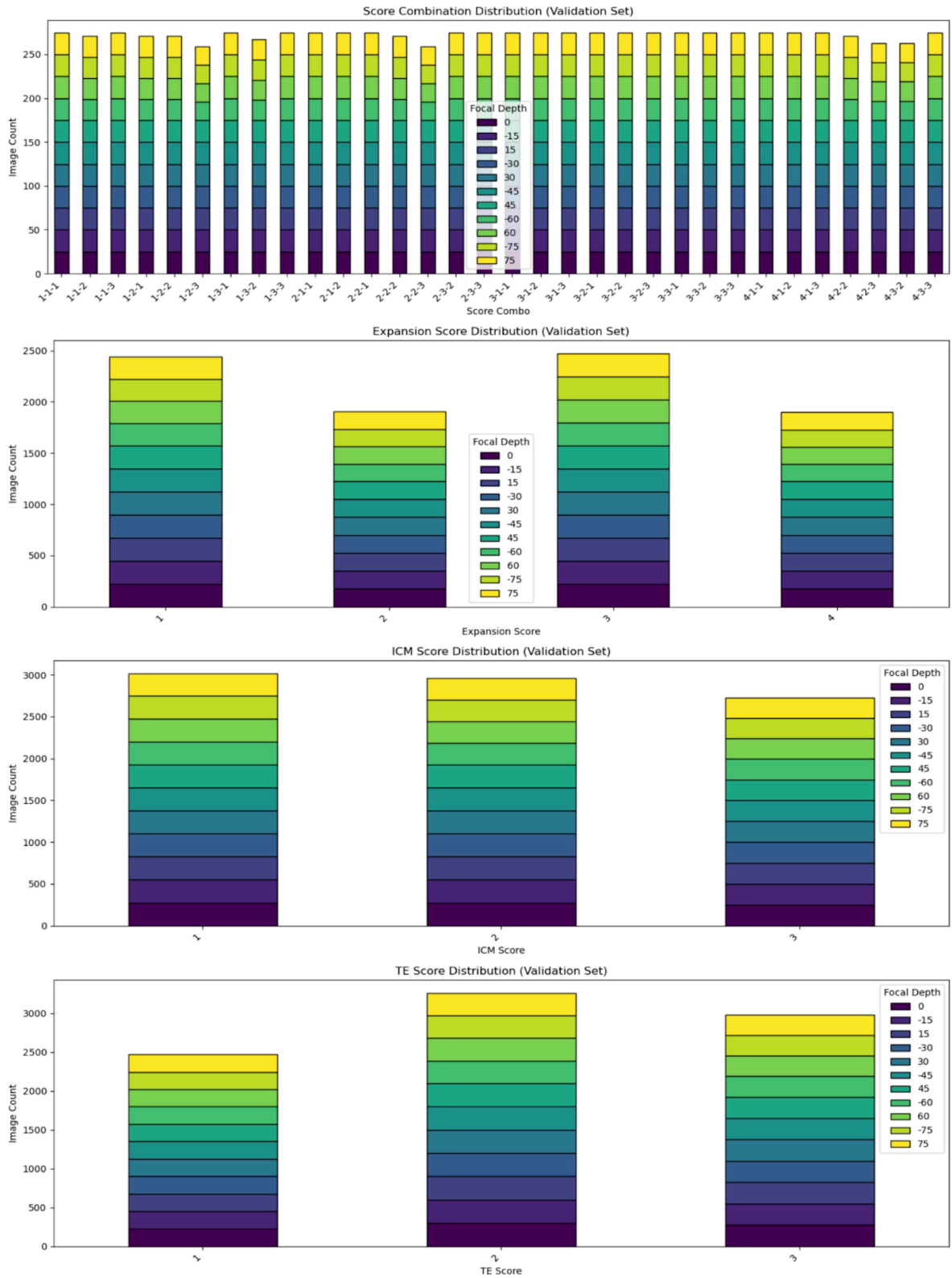

**Supplemental Figure 7:** DIA validation data distribution across cumulative scores and individual scoring categories. Score combo in the first graph refers to (Expansion-ICM-TE)

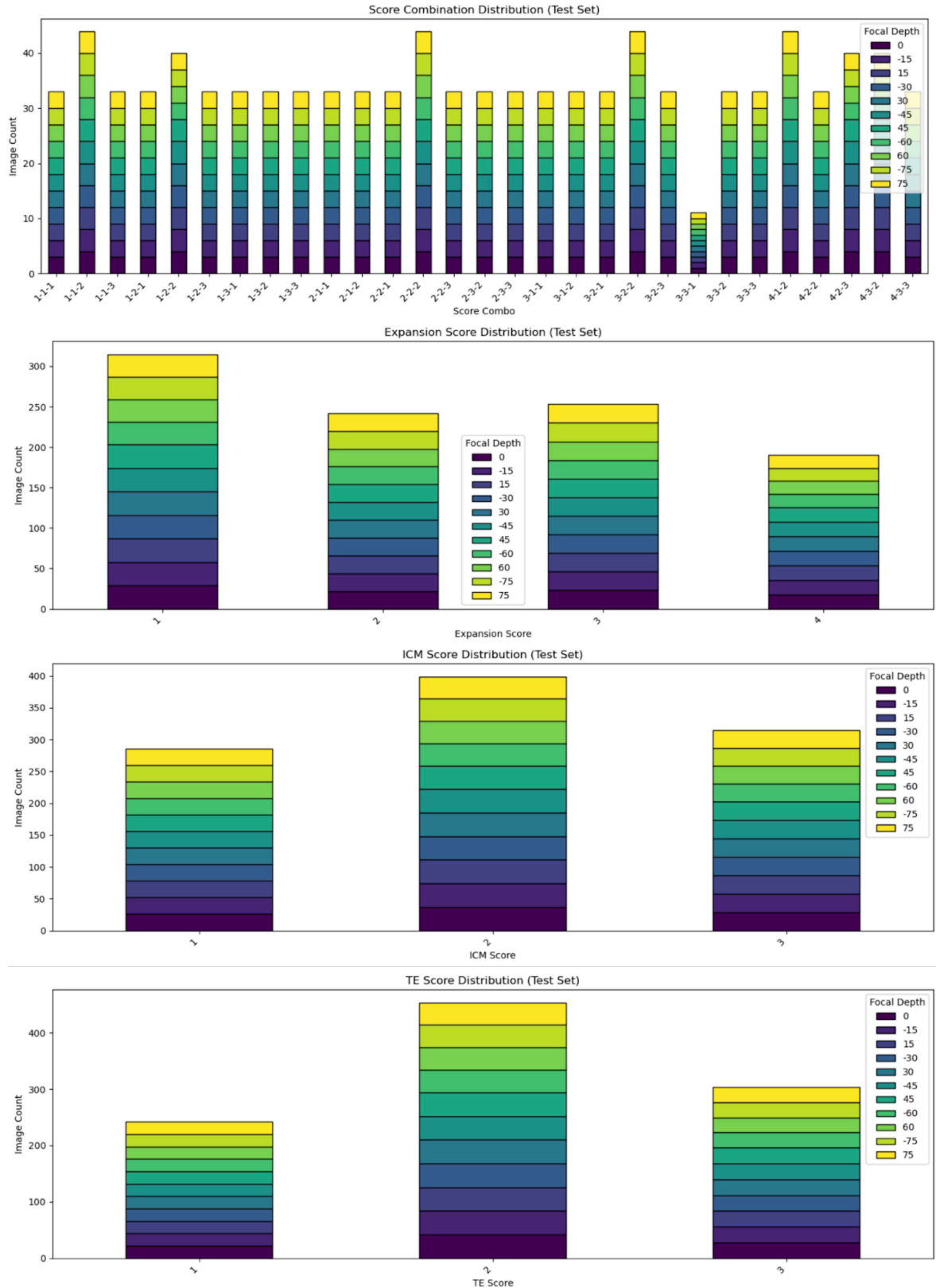

**Supplemental Figure 8:** DIA test data distribution across cumulative scores and individual scoring categories. Score combo in the first graph refers to (Expansion-ICM-TE)

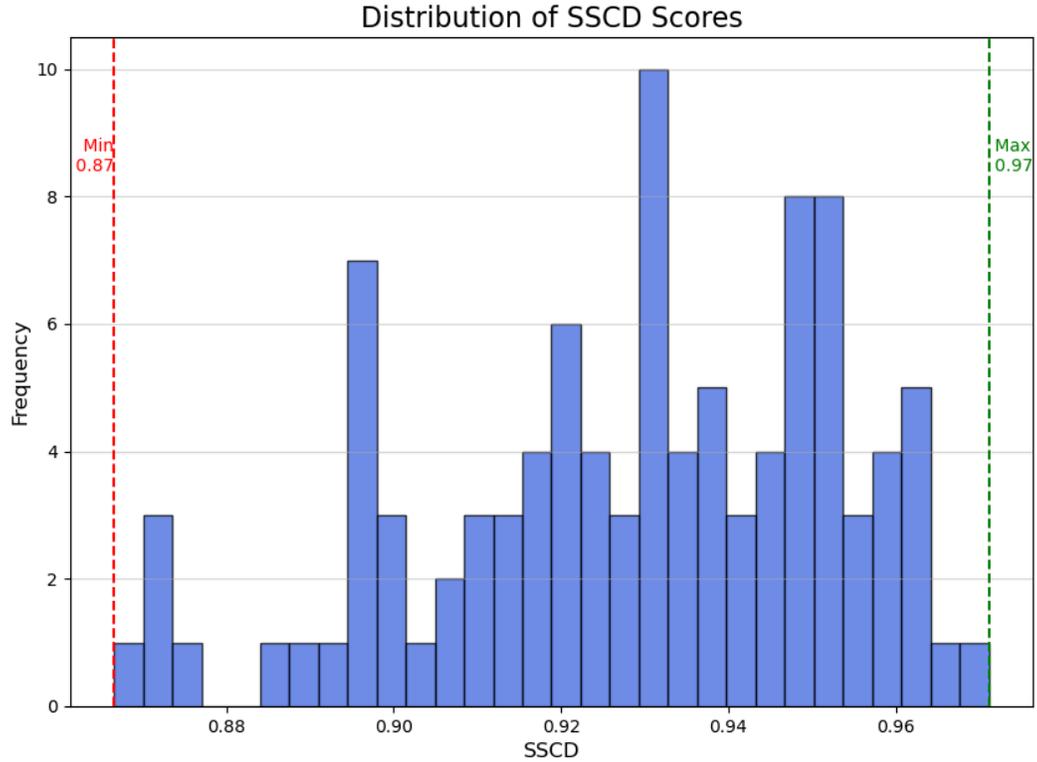

**Supplemental Figure 9:** SSCD scores for real images from LDM dataset augmented between -10,+10 pixels in X and Y direction and rotated between -5, 5 degrees and all images in the training set. The highest SSCD score of the augmented image corresponded with the original non-augmented image 99/100 times.

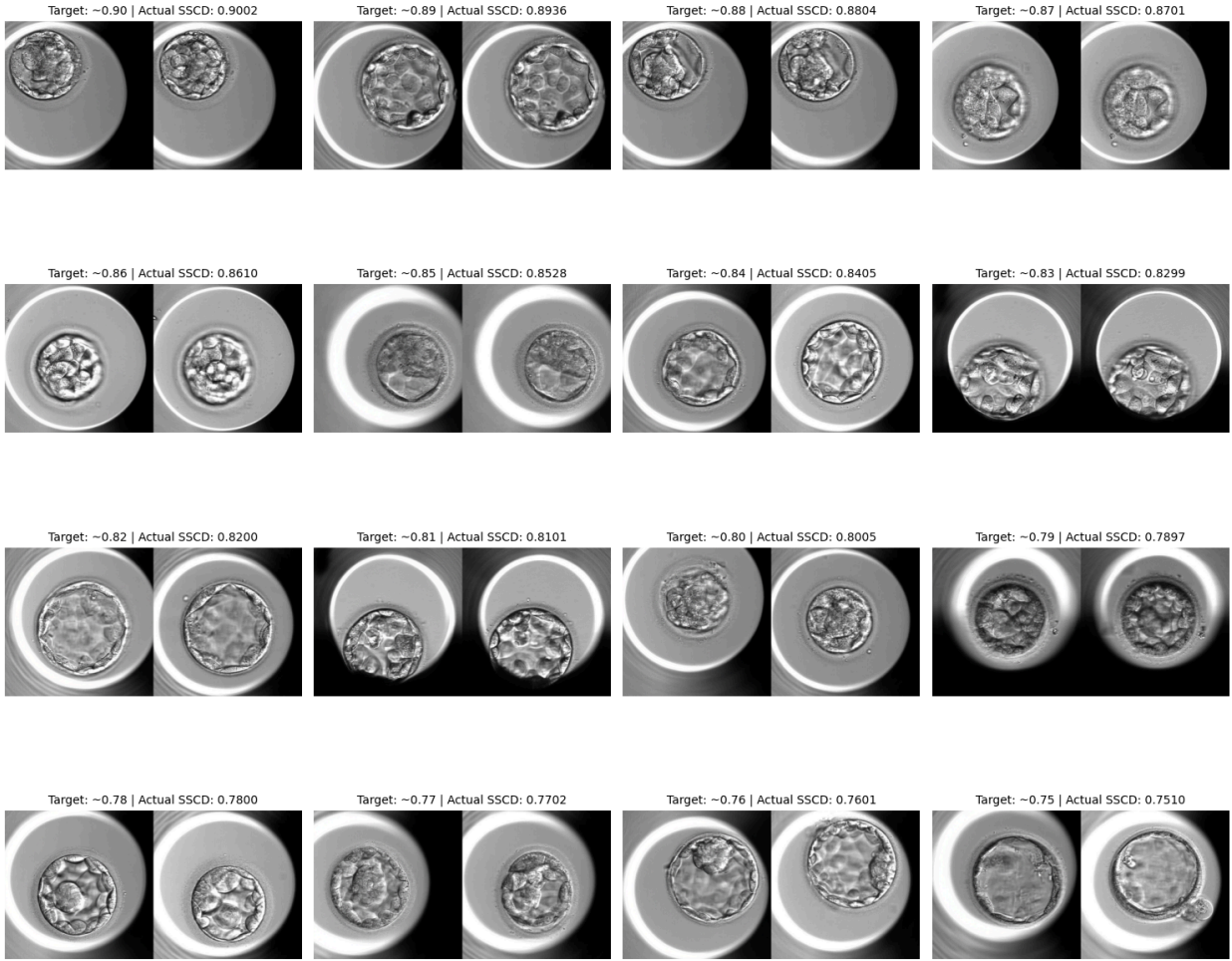

**Supplemental Figure 10:** Synthetic and closest real images from training dataset at various SSCD values.

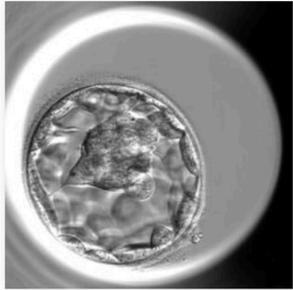

**Supplemental Figure 11:** Web portal of the Turing Test

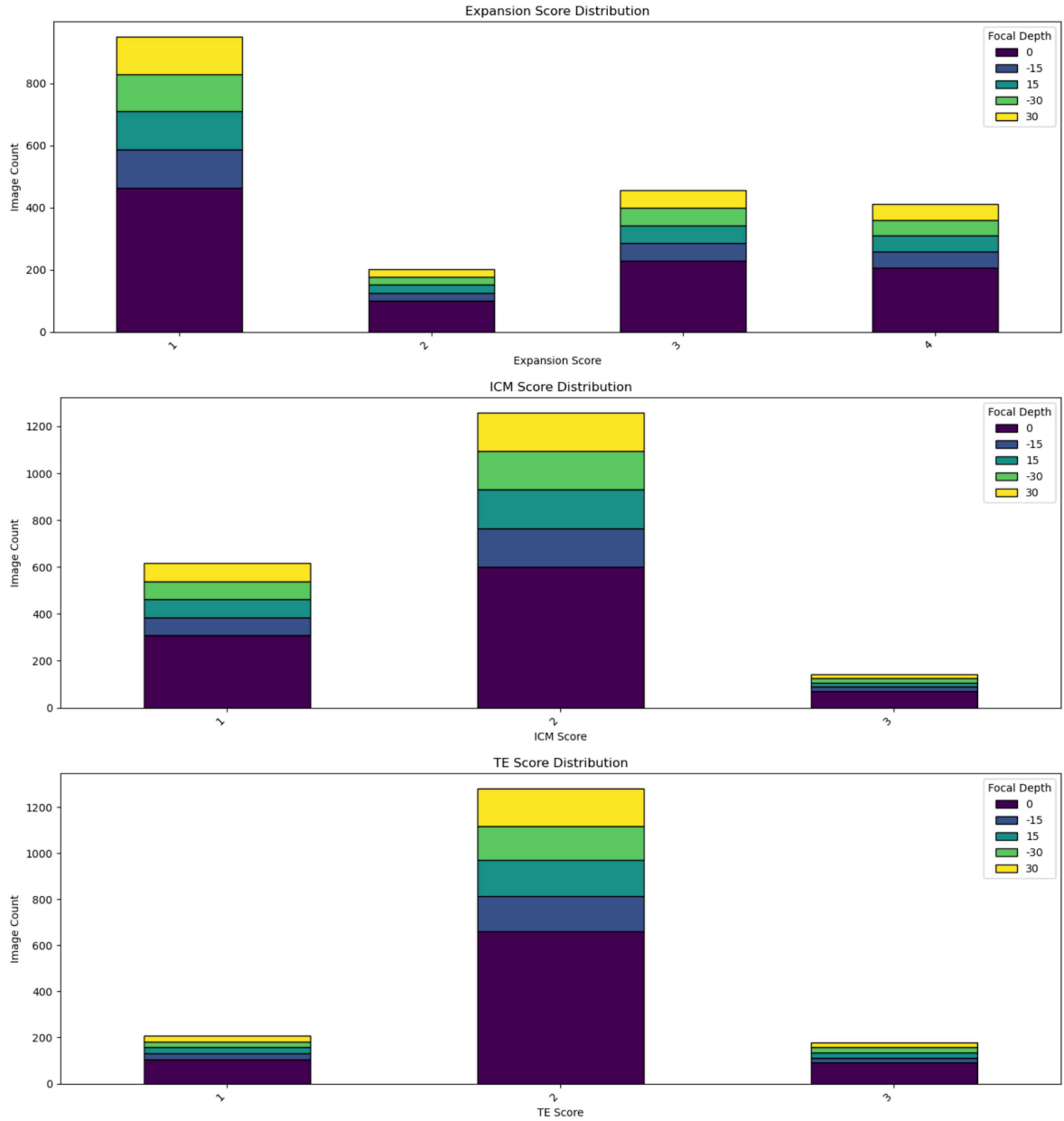

**Supplemental Figure 12:** Real world distribution baseline ResNet dataset (sample training fold data shown)

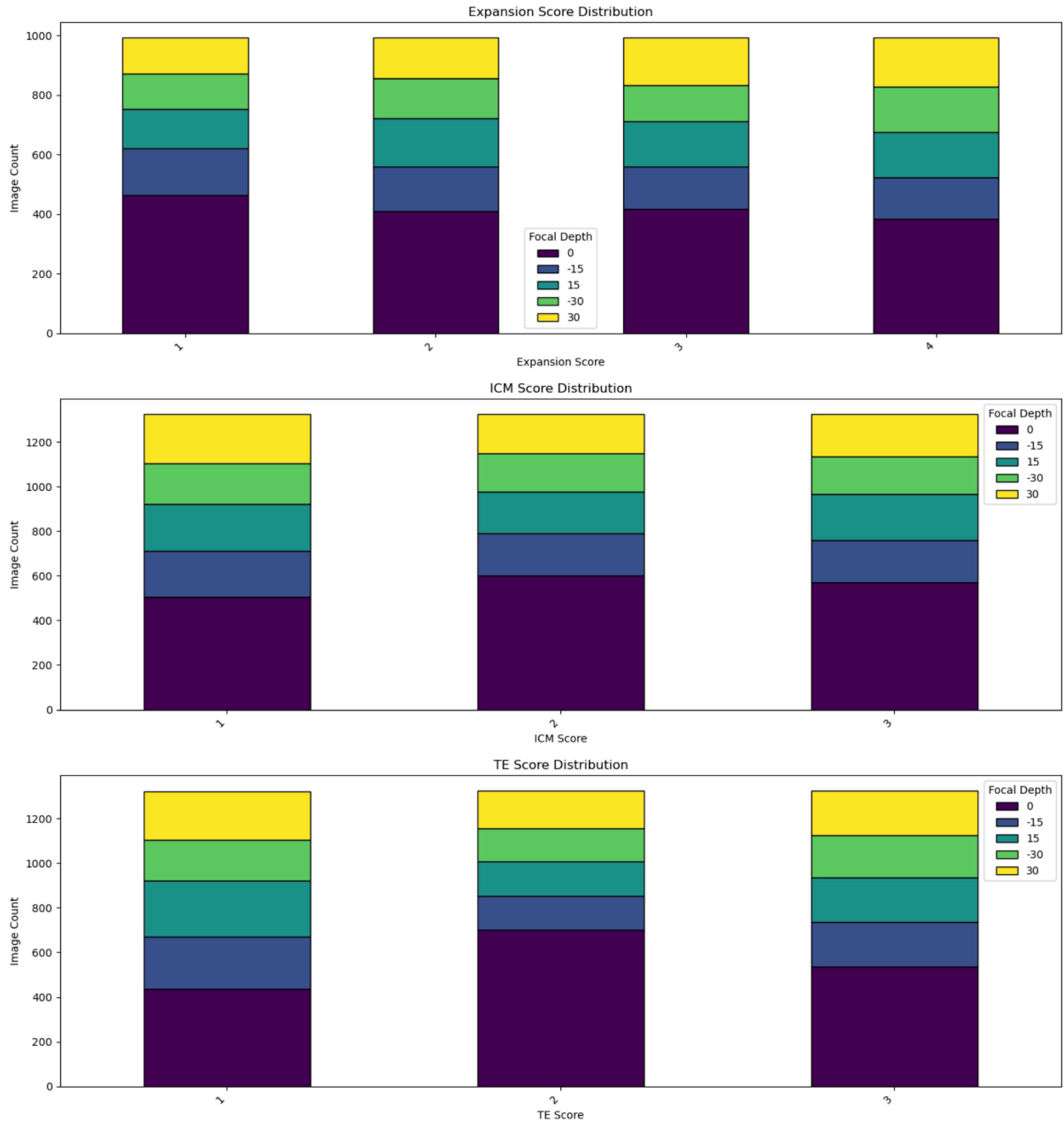

**Supplemental Figure 13:** Real image only balanced baseline ResNet dataset (sample training fold data shown)

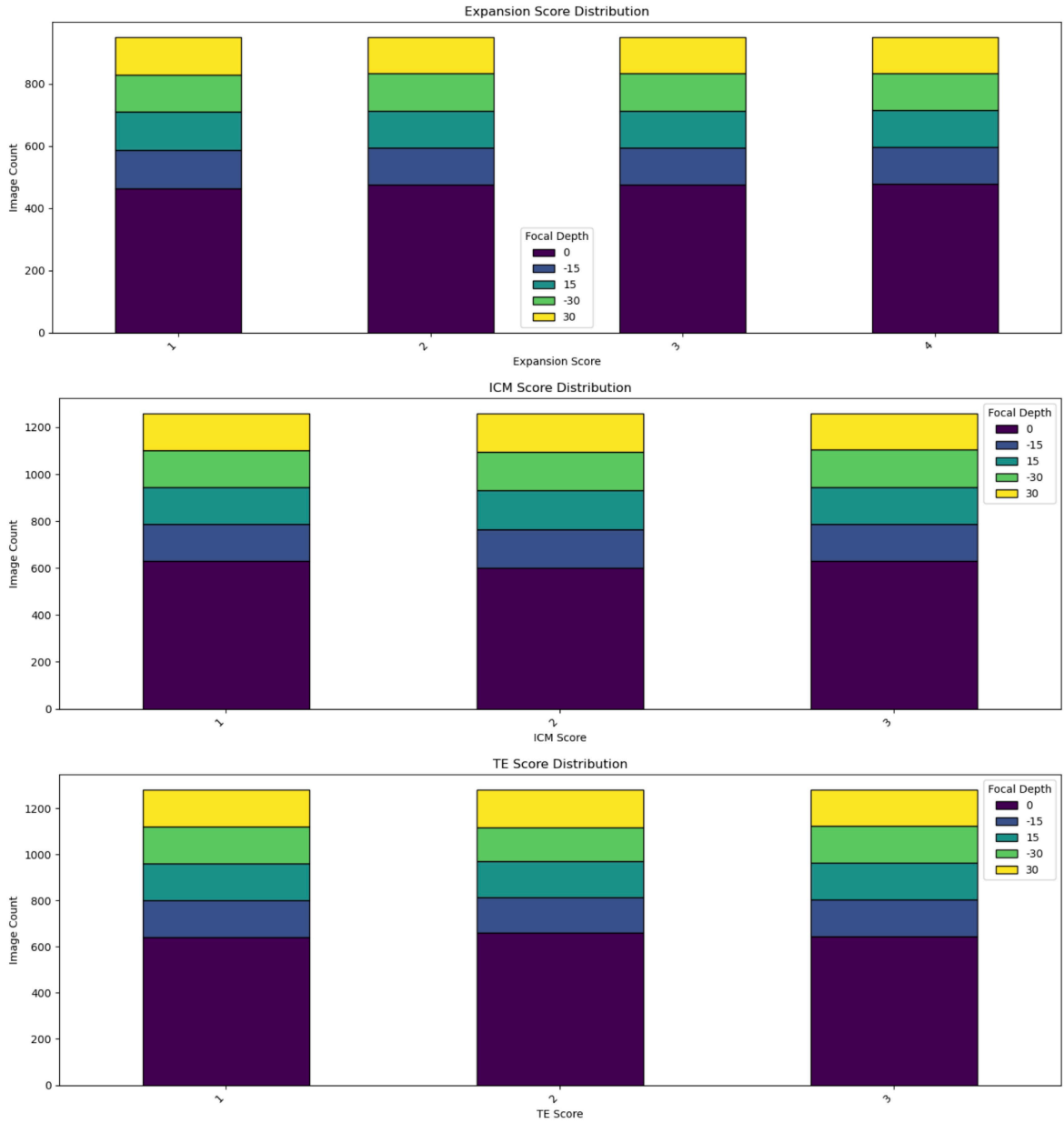

**Supplemental Figure 14:** Class balance distribution of images once imbalanced baseline augmented with synthetic images (sample training fold data shown)